\documentclass{article}
\usepackage{amd}
\usepackage[utf8]{inputenc}
\usepackage[T1]{fontenc}
\usepackage{microtype}
\usepackage{amsmath,amssymb,amsthm,booktabs,tabularx}
\usepackage{flafter}
\usepackage[section]{placeins}
\AddToHook{env/table/begin}{\small\renewcommand{\arraystretch}{1.12}}
\newcommand{\estci}[3]{$#1\;[#2,#3]$}
\usepackage{tikz}
\usepackage{graphicx}
\usepackage{pgfplots}
\pgfplotsset{compat=1.18}
\usetikzlibrary{arrows.meta,positioning,fit,calc,patterns}
\DeclareMathSizes{7.2}{7.2}{5}{5}
\DeclareMathSizes{7.3}{7.3}{5}{5}
\definecolor{FigInk}{HTML}{181818}
\definecolor{FigMuted}{HTML}{4B4B4B}
\definecolor{FigGrid}{HTML}{D7D7D7}
\definecolor{FigBlue}{HTML}{8EB9DA}
\definecolor{FigTeal}{HTML}{A6D5C4}
\definecolor{FigAmber}{HTML}{E9C698}
\definecolor{FigRose}{HTML}{DBA5AE}
\definecolor{FigGray}{HTML}{BFC4C7}
\definecolor{FigPanel}{HTML}{F5F5F3}
\definecolor{FigBlueInk}{HTML}{0072B2}
\definecolor{FigTealInk}{HTML}{007F60}
\tikzset{
  fig/base/.style={font=\rmfamily\fontsize{8}{9.5}\selectfont,
    text=FigInk,line cap=round,line join=round},
  fig/title/.style={font=\rmfamily\fontsize{9}{10.5}\selectfont\bfseries,
    text=FigInk,anchor=west,inner sep=0pt},
  fig/small/.style={font=\rmfamily\fontsize{8}{9.5}\selectfont,text=FigMuted},
  fig/arrow/.style={-{Latex[length=1.45mm,width=.95mm]},
    draw=FigInk,line width=.5pt},
  fig/transfer/.style={fig/arrow,dashed},
  fig/panel/.style={draw=FigInk,fill=FigPanel,line width=.45pt,
    rounded corners=2pt},
  fig/box/.style={draw=FigInk,fill=white,line width=.45pt,
    rounded corners=.6pt,align=center,inner xsep=4pt,inner ysep=4pt},
  fig/teacher/.style={fig/box,fill=FigBlue!45},
  fig/learned/.style={fig/box,fill=FigTeal!55},
  fig/axis/.style={draw=FigInk,line width=.5pt},
  fig/grid/.style={draw=FigGrid,line width=.35pt,densely dotted},
  fig/value/.style={font=\rmfamily\fontsize{8}{9.5}\selectfont,
    text=FigInk,inner sep=1pt}
}

\definecolor{TableGain}{HTML}{86BC91}
\newcommand{\gain}[2]{\begingroup\setlength{\fboxsep}{1.2pt}\colorbox{TableGain!#1}{\strut#2}\hspace{-1.2pt}\endgroup}
\usepackage{hyperref}
\hypersetup{hidelinks}
\usepackage{url}

\newcommand{\KL}{\operatorname{KL}}
\newcommand{\E}{\mathbb{E}}
\newcommand{\Var}{\operatorname{Var}}

\newcommand{\softmax}{\operatorname{softmax}}
\IfFileExists{assets/AMD_logo.png}{\newcommand{\AMDlogo}{\includegraphics[height=0.28in]{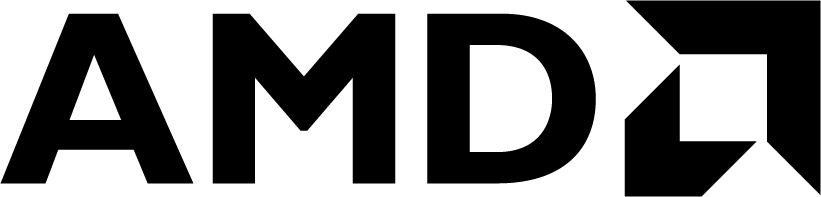}}}{}
\hypersetup{pdftitle={TempQuant: Learning Adaptive Temperatures for Quantized Language Models},pdfauthor={Mukund Athmaram, Liang Ma, Manoj Rao}}
\begin{document}
\onecolumn
\amdtitlerunning{TempQuant: Learning Adaptive Temperatures for Quantized Language Models}
\amdtitle{TempQuant: Learning Adaptive Temperatures for Quantized Language Models}

\begin{amdauthorlist}
  \amdauthor{Mukund Athmaram}{amd,cornell,cor}
  \amdauthor{Liang Ma}{amd}
  \amdauthor{Manoj Rao}{amd}
\end{amdauthorlist}
\amdaffiliation{amd}{AMD}
\amdaffiliation{cornell}{Cornell University}
\amdcorrespondingauthor{Mukund Athmaram}{ma2388@cornell.edu}
\amdkeywords{quantization, sampling temperature, top-p, large language models, decoding}
\vskip 0.3in
\amdPrintAffiliationsInline
{\let\thefootnote\relax\footnotetext{Work conducted during an AMD internship.}}

% === Abstract ===
\begin{abstract}
Large language models are increasingly deployed at low precision for
applications ranging from reasoning to creative writing, yet quantization
can degrade their performance. These applications often use different
decoding settings, such as the sampling temperature and top-$p$, which
control the randomness of token selection. We study how quantization
interacts with these settings by examining how it distorts the model's
next-token probabilities, which can change generated responses and task
performance. Measuring entropy during generation across deployed
temperatures, we find that quantization warms the model on average, and that
temperature matters much more for the performance of a quantized model than
for its full-precision counterpart. Because quantization and temperature
both shape the next-token distribution, we investigate how much the sampling
settings can undo what quantization changes, through our method TempQuant,
which learns the quantized model's temperature and top-$p$ to bring its
next-token probabilities closer to the full-precision model's. We learn
these settings at two granularities. The first is a single global setting
per requested temperature, which we show finds nearly the same temperatures
that tuning on labeled answers would, without any labels. This matters in
practice because model cards typically recommend sampling a quantized model
with the same settings as its full-precision counterpart, while we find that
different settings can substantially improve its accuracy. On 3-bit
Qwen3-4B, for example, global TempQuant raises MATH accuracy from $34\%$
under the configuration recommended for Qwen's quantized releases to
$51\%$. The second sets the
temperature and top-$p$ for every token. It is motivated by our finding that
tokens whose rankings flip under quantization tend to need warming, while tokens
whose rankings are preserved tend to need cooling, which one temperature cannot do
at once. Finally, we show that when quantization degrades accuracy
substantially, as at 3 bits, TempQuant recovers much of it, and when
degradation is lighter and there is little accuracy to recover, it still
brings the quantized model's behavior closer to the full-precision model's.
\end{abstract}

% === Introduction ===
\section{Introduction}
\label{sec:introduction}

At every step of autoregressive generation, large language models generate
a probability distribution over the next token. This probability distribution
is a function of the prefix, which is the prompt
together with any tokens generated so far. Based on the chosen decoding
method, we select a token from this distribution \citep{shi2024decoding}.
When sampling, the randomness of token selection is controlled by the
temperature $\tau$, a user-chosen deployment setting. The chosen temperature
depends on the use case: higher temperatures generally lead to more diverse
outputs, whereas lower temperatures concentrate sampling on the most likely
tokens. Changing the temperature changes the entropy of the distribution, that is,
how spread out it is.

Quantization is a separate form of
compression, where we use fewer bits to represent the weights and, in some
configurations, activations of the large language model, so it can fit in less
memory and run faster on supporting hardware \citep{frantar2023gptq,liu2023llmqat}.
Compressing this information introduces numerical error, so the next-token
probability distributions can diverge from those of the full-precision model.
These distortions can also change the distribution's entropy, affecting the
sampling randomness that the user controls through temperature.

Temperature scaling and quantization thus both change the model's
next-token distribution, and we analyze how the two interact. How well a
decoding setting works depends on the task and on factors such as
quantization, and the best results often require extensive tuning
\citep{shi2024decoding}, which motivates learning decoding settings
automatically \citep{wang2025autodeco}. We study how much of quantization's
distortion of the next-token distribution can be corrected by adjusting
temperature and top-$p$, and whether doing so recovers accuracy.

We quantify the difference between the full-precision and quantized
probability distributions through Kullback--Leibler divergence
\citep{kullback1951information}. We hypothesize that reducing this divergence
can help preserve the full-precision model's behavior. This is motivated by
\emph{Accuracy is Not All You Need}, in which \citet{dutta2024accuracy} argue that quantization should preserve the
full-precision model's behavior, and that similar aggregate benchmark
accuracy does not establish behavioral preservation, because
correct-to-incorrect and incorrect-to-correct ``flips'' can cancel out.
They motivate distance metrics such as KL divergence to capture the changes
that aggregate accuracy hides, and this divergence is what we reduce.

Minimizing divergence between probability distributions is not new in quantization:
LLM-QAT updates quantized model weights to minimize forward KL for fixed
teacher targets \citep{liu2023llmqat}. However, quantization-aware training (QAT) retrains the language-model
weights, which can be expensive. We instead reduce this divergence with a
lightweight method applied after post-training quantization (PTQ). We modify two sampling settings, temperature and top-$p$, rather than the
model weights, and call this method TempQuant. We learn these settings at two
granularities: global TempQuant learns one temperature and top-$p$ to deploy the quantized
model at for each temperature requested of the full-precision model, and
per-step TempQuant chooses them at every decoding step. Global TempQuant is the practical version that can be applied today, since
it only changes the decoding settings. We also show why changing the
temperature at every token can recover even more of the quantization error,
which motivates per-step TempQuant.

\begin{figure}[!tb]
\centering
\includegraphics[width=\linewidth]{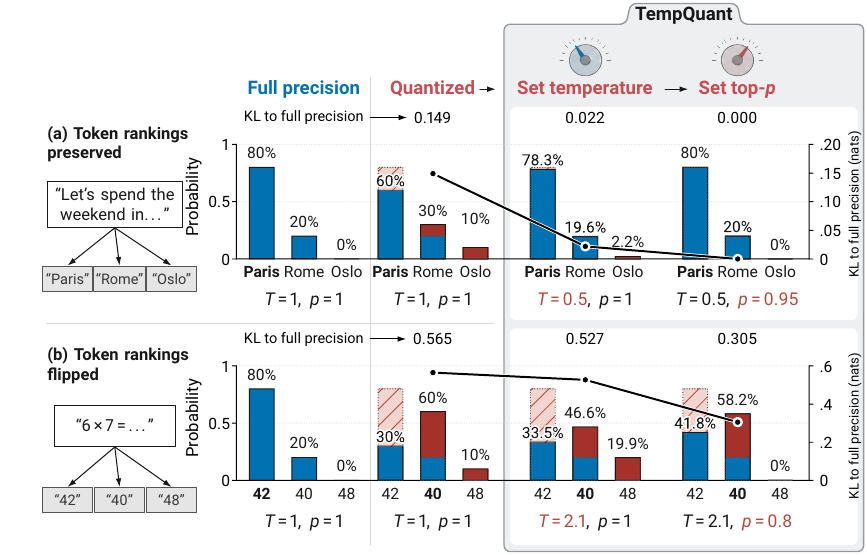}
\caption{Illustrative three-token distributions before and after temperature and top-$p$ correction.}
\label{fig:temperature-mechanism}
\end{figure}

Figure~\ref{fig:temperature-mechanism} illustrates how decoding settings
change token probabilities. At every token there is a best temperature and
top-$p$, and per-step TempQuant aims to choose it.

In practice, existing deployments usually sample a quantized model at the
settings chosen for its full-precision counterpart. Qwen, for
example, recommends temperature $0.6$ and top-$p$ $0.95$ for the
full-precision Qwen3 \citep{qwen3card4b} and repeats the same settings
for its own 4-bit and GGUF releases \citep{qwen3card4bawq,qwen3card8bgguf}, and community
quantizations down to 3 bits copy these settings unchanged
\citep{ollamaqwen3,unslothqwen3gguf}. We argue that this leaves much of the quantization error unrecovered, and
that choosing different sampling settings for the quantized model, whether
at a global or a per-token granularity, can recover much of it
(Section~\ref{sec:downstream}).

% === Related Work ===
\section{Related Work}
\label{sec:related}

\paragraph{Adaptive decoding.}
Several methods learn temperatures at a fine granularity: ATS per token for
calibration \citep{xie2024ats}, TempNet per input for task performance
\citep{qiu2024tempnet}, and ReSET per token for quantized reasoning models,
with settings tuned on labeled data \citep{lee2026reset}. AutoDeco predicts
temperature and top-$p$ at every step with two small heads
\citep{wang2025autodeco}, a design we borrow, but trains them to raise the
likelihood of reasoning traces that reach correct answers, so as to replace
hand-tuned decoding. TempQuant instead learns its settings to match the
full-precision model, so as to undo the effect of quantization.

\paragraph{Decoding-time fixes.}
\citet{lotfi2026overthinking} measure the token-level KL between quantized
and full-precision reasoning models and reduce overthinking with a logit
penalty on a curated set of markers.
TempQuant also corrects the quantized model at decoding time, but fits the
standard temperature and top-$p$ to the full-precision model. Other samplers
target repetition without a reference model
\citep{keskar2019ctrl,basu2021mirostat,meister2023typical,hewitt2022truncation,nguyen2025minp},
and \citet{braverman2020calibration} correct entropy-rate drift with a single
temperature-like parameter, a close precedent for global TempQuant.

\paragraph{Temperature calibration and quantization.}
DACA, Thermometer, and Dual-Align fit temperatures for confidence
calibration \citep{luo2025daca,shen2024thermometer,luo2026dualalign}, while
we fit them to keep a quantized model close to its full-precision version.
Aggressive quantization can lengthen chains of thought and cause repetition
\citep{lotfi2026overthinking,li2024evaluatingquantized,liu2025quantizationhurts},
and can flip preferences between candidate tokens \citep{wu2026decisions}.
We use GPTQ \citep{frantar2023gptq}, AWQ \citep{lin2024awq}, and the MXFP4
format \citep{rouhani2023microscaling}. Appendix~\ref{app:related} discusses
related work in more detail.

% === Method ===
\section{Method}
\label{sec:final-method}

\paragraph{Matching the full-precision distribution.}
\label{sec:recoverability}
Our goal is to bring the probability distribution the quantized model emits
at each step as close as possible to what the full-precision model emits
under the user's requested temperature and top-$p$. We compare the two models after the same prefix. Temperature divides the
logits before they are turned into probabilities, and top-$p$ keeps the most probable tokens until their
cumulative probability reaches the cutoff, then rescales them to sum to one
\citep{holtzman2020degeneration}.

Let $u_t,z_t\in\mathbb{R}^{V}$ be the full-precision and quantized logits at
prefix $c_t$, and let $\mathcal N_r$ denote top-$p$ filtering with cutoff $r$.
With requested temperature $\tau>0$ and cutoff $\rho$, the full-precision
model emits
\begin{equation}
 p_t=\mathcal N_\rho(\softmax\!\left(\frac{u_t}{\tau}\right)),
 \label{eq:full-precision-distribution}
\end{equation}
and we fix $\rho=.95$ in our experiments. If we instead sample the quantized
model with temperature $T_t$ and cutoff $\rho_t^Q$, it emits
\begin{equation}
 q_t=\mathcal N_{\rho_t^Q}(\softmax\!\left(\frac{z_t}{T_t}\right)).
 \label{eq:final-distributions}
\end{equation}
We learn $T_t$ and $\rho_t^Q$ so that $q_t$ is close to $p_t$. Both forms of TempQuant are learned against the full-precision model on the
same training data, and both need only the quantized model at deployment. We measure how close $q_t$ is to $p_t$ with forward
KL, $\KL(p_t\Vert q_t)$, reverse KL, $\KL(q_t\Vert p_t)$, or
$\mathrm{JSD}(p_t,q_t)$, and Section~\ref{sec:divergence} explains why we use all three. Both KL losses mix each distribution with a uniform
component of weight $\eta=10^{-4}$ so they stay finite when top-$p$ removes
a token the other model keeps (Appendix~\ref{app:smoothed-divergences}).
Unless noted, divergences in this paper use these smoothed KLs between
top-$p$-filtered distributions.

\paragraph{Global TempQuant.}
For each requested temperature, global TempQuant searches a grid of temperature and top-$p$
values for the pair with the lowest divergence on the training data
(Appendix~\ref{app:learned-global}). It is fit separately for each quantized model, and is temperature scaling \citep{guo2017calibration} with the full-precision
model's next-token distribution in place of the labels, together with
top-$p$. We also learn two
smaller versions: global TempQuant (top-$p$ only), which keeps the requested
temperature, and global TempQuant ($T$ only), which keeps $\rho_t^Q=.95$. Global TempQuant adds no cost at deployment, since it only
changes the two numbers passed to the sampler.

\paragraph{Per-step TempQuant.}
Per-step TempQuant chooses the setting at every step. At inference we no longer have the full-precision
model, so it has to choose from what the quantized model gives us. At each
step, it looks at the quantized logits after adjusting them for the
requested temperature, and summarizes them in 42 features: the 32 largest
logits, normalized, and ten statistics that describe how spread out the
distribution is, such as its entropy and the gaps between the leading
tokens. Two small MLP heads read these features and predict a correction
$\kappa_t$ to the requested temperature and a new top-$p$ cutoff. The
sampler divides the original quantized logits by $T_t=\tau\kappa_t$ and
applies the predicted cutoff. We borrow this two-head design from AutoDeco
\citep{wang2025autodeco}. With $x_t(\tau)$ denoting the features,
\begin{equation}
 \kappa_t=F_\theta(x_t(\tau)),\qquad
 \rho_t^Q=G_\phi(x_t(\tau),\log\kappa_t),\qquad
 T_t=\tau\kappa_t.
 \label{eq:final-heads}
\end{equation}
Appendix Figure~\ref{fig:inference} shows how the heads control the sampler. We also
train per-step TempQuant ($T$ only), which only predicts the temperature and keeps
$\rho_t^Q=.95$. Where a table lists training losses, ``TempQuant''
refers to per-step TempQuant.
Appendix~\ref{app:final-spec} gives the exact features, output ranges, and
training details.

\paragraph{Training the per-step predictor.}
Both language models remain frozen, and we train only the two heads,
comparing full next-token distributions at every calibration prefix as in
LLM-QAT \citep{liu2023llmqat}, with the sampler's settings in place of the
weights. The
calibration prefixes come from responses by both BF16 and the quantized
model (Appendix~\ref{app:training-data}), so the loss covers text from both
the reference and the quantized model. Each prefix is evaluated at
requested temperatures $0.3$, $0.6$, and $1.0$, with all prefix and
temperature pairs weighted equally. Hard top-$p$ gives no gradient for the
cutoff, so training uses a smooth version of the quantized model's top-$p$,
as in AutoDeco, while the BF16 reference keeps hard filtering. We minimize
\begin{equation}
 \mathcal L_D=\frac1M\sum_{j=1}^{M} D\big(p_j,\widetilde q_j\big)
 +\lambda_D\,\mathcal R,
 \label{eq:final-objective}
\end{equation}
where $j$ runs over all $M$ pairs of a training prefix and a requested
temperature, $D$ is one of the three divergences, and $\widetilde q_j$ uses
the smooth top-$p$. The penalty $\mathcal R$ measures how much the
predicted temperature and top-$p$ vary from one training prefix to the
next. We compute it separately at each requested temperature, $0.3$, $0.6$, and
$1.0$, and average the three. It keeps the heads from overfitting the small
calibration set, and it also pulls them toward a single setting
(Appendix~\ref{app:final-spec}).

% === Experiments ===
\section{Experiments}
\label{sec:final-evaluation}
\label{sec:final-analysis}
We first analyze how quantization changes the next-token distribution
(Sections~\ref{sec:temperature-results}--\ref{sec:divergence}), then
evaluate both forms of TempQuant
(Sections~\ref{sec:adaptive-results}--\ref{sec:per-step}), their accuracy
(Section~\ref{sec:downstream}), and how well they preserve the
full-precision model's behavior (Section~\ref{sec:behavior-results}).

We evaluate Qwen3-4B \citep{qwen3technicalreport} quantized to 3-bit GPTQ
(GPTQ3) and to native 4-bit MXFP4, at requested temperatures $0.3$, $0.6$,
the setting Qwen recommends, and $1.0$, with further models in
Section~\ref{sec:downstream}. Appendix~\ref{app:campaign-protocol}
gives the benchmarks, data splits, and statistical tests.

\subsection{How quantization changes the temperature}
\label{sec:temperature-results}
\begin{figure}[!htb]
\centering
\includegraphics[width=\linewidth]{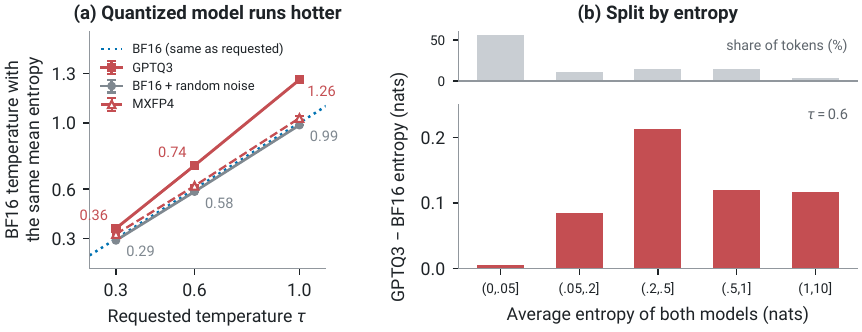}
\caption{Quantized models against BF16: (a) the BF16 temperature with the same mean entropy as GPTQ3, MXFP4, and BF16 with random noise added; (b) GPTQ3's entropy change from BF16 at $\tau=0.6$, grouped by the average entropy of the two models, with each group's share of tokens above.}
\label{fig:three-bit-context}
\end{figure}

We find that quantized models have higher average entropy than BF16: their
next-token probabilities are more spread out, as if they were sampled at a
higher temperature. GPTQ3 at a requested $0.6$ has the same mean entropy as BF16 at
$0.74$, and the gap grows with the requested temperature
(Figure~\ref{fig:three-bit-context}a). Adding random noise of the same size to BF16's logits leaves its effective
temperature where it was, so the heating comes from quantization itself
(Appendix~\ref{app:noise-temperature}). MXFP4, which loses
much less accuracy, shows a smaller shift in the same direction
(Figure~\ref{fig:three-bit-context}a).

Quantization barely changes the near-certain tokens that make up most of
the text, and the extra entropy comes from tokens where the
model is more uncertain (Figure~\ref{fig:three-bit-context}b).

\subsection{Which tokens carry the quantization error}
\label{sec:which-way}
\begin{figure}[!tb]
\centering
\includegraphics[width=\linewidth]{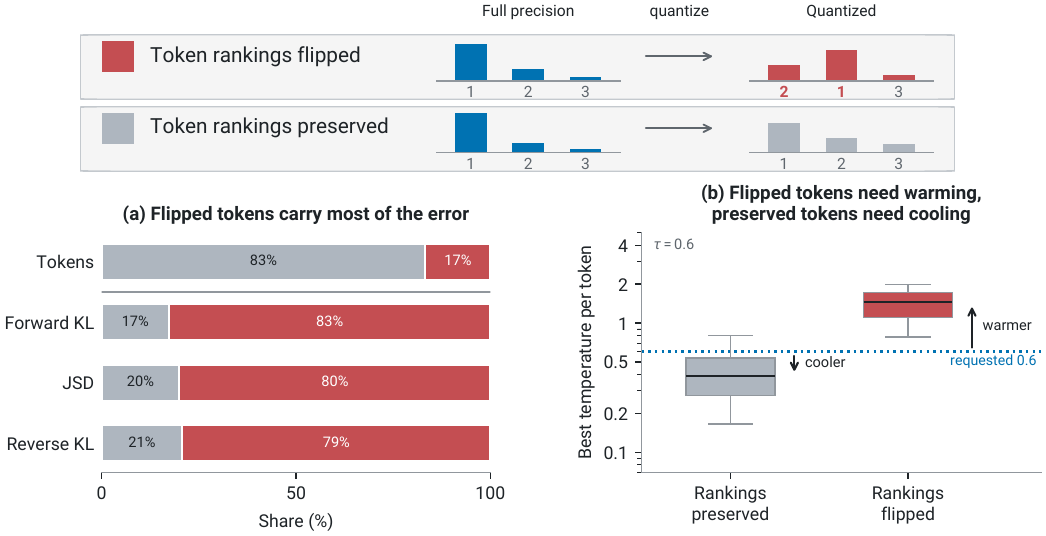}
\caption{GPTQ3 against BF16 at $\tau=0.6$, on tokens where the rankings are preserved or flipped: (a) share of tokens and of each divergence; (b) the temperature that best matches BF16 for each token under forward KL.}
\label{fig:which-way}
\end{figure}
Since the quantized model runs hotter, it is natural to assume that cooling
it would reduce this divergence. However, our proxy for quantization error
is KL divergence, not entropy, and whether warming or cooling reduces KL
divergence depends on what quantization does to the token probabilities.
Quantization does two different things to tokens
(Figure~\ref{fig:which-way}). We say a token's rankings are preserved when
the quantized model's most likely next token is the same as BF16's, and
flipped when it is not; less likely tokens may still change order. On most
tokens, quantization keeps BF16's most likely token and only spreads the
probabilities out. On these tokens, where the rankings are preserved,
quantization acts like heating, so they tend to be best
matched by cooling to a lower temperature. On the rest, however, the
rankings flip and BF16's preferred token now ranks lower. Here only a higher
temperature gives it back probability, as in
Figure~\ref{fig:temperature-mechanism}b, so raising the temperature tends
to reduce the KL divergence (Figure~\ref{fig:which-way}b). These flipped
tokens are a small share of tokens, but each costs far more, so they carry
most of the error under every divergence we use
(Figure~\ref{fig:which-way}a), at every requested temperature (Appendix
Table~\ref{tab:flip-share}).

A single temperature therefore has to trade off the two, since warming helps
the flipped tokens and cooling helps the preserved ones. Fitting global
TempQuant only on preserved tokens shows this. Its temperature drops close
to the requested one and fits those tokens slightly better, but the error
rises on flipped tokens and over all tokens
(Appendix~\ref{app:learned-global}). A per-step temperature can instead
correct each token on its own.

\subsection{Choice of divergence}
\label{sec:divergence}
To see how handling this trade-off differently changes performance, we
measure the quantization error with three divergences that weigh the two
kinds of token differently. TempQuant can be seen as a restricted form of
knowledge distillation \citep{hinton2015distilling}, in which the
full-precision model is the teacher, its quantized version is the student,
and the student can lower the divergence only through its sampling
settings, without changing its weights. On-policy distillation
\citep{agarwal2024gkd} shows that when a student cannot reproduce its
teacher, the divergence decides which errors it accepts. Inspired by this,
we train with three divergences that fail in different ways
(Appendix~\ref{app:divergences}).

Forward KL is large when the quantized model gives little probability to a
token BF16 considers likely. A flipped token therefore spikes forward KL, so forward KL prioritizes
warming to correct the flipped tokens. Reverse KL, on the other hand, is large when the
quantized model puts probability where BF16 does not. Warming spreads
probability onto tokens BF16 gives little probability to, which spikes
reverse KL, so reverse KL prefers cooling even for the flipped tokens. JSD combines the
two, weighs both errors, and lands in between, near the requested
temperature. On a typical preserved token every divergence favors cooling,
so they part ways mainly on the flipped tokens, as Appendix~\ref{app:divergence-toy}
illustrates. This is the familiar contrast between mass-covering and
mode-seeking divergences.

\begin{figure}[!htb]
\centering
\includegraphics[width=\linewidth]{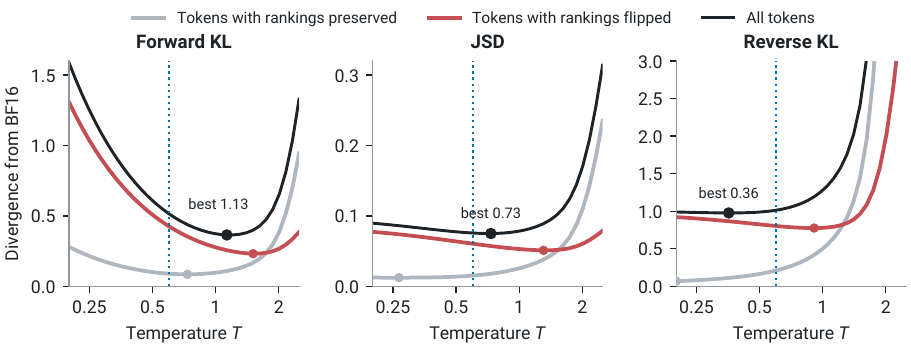}
\caption{Divergence from BF16 at $\tau=0.6$ when one temperature is used for every token, split into the part from tokens with preserved and flipped rankings. Dots mark each curve's minimum, and the dotted line the requested temperature.}
\label{fig:tradeoff}
\end{figure}

\begin{figure}[!htb]
\centering
\includegraphics[width=\linewidth]{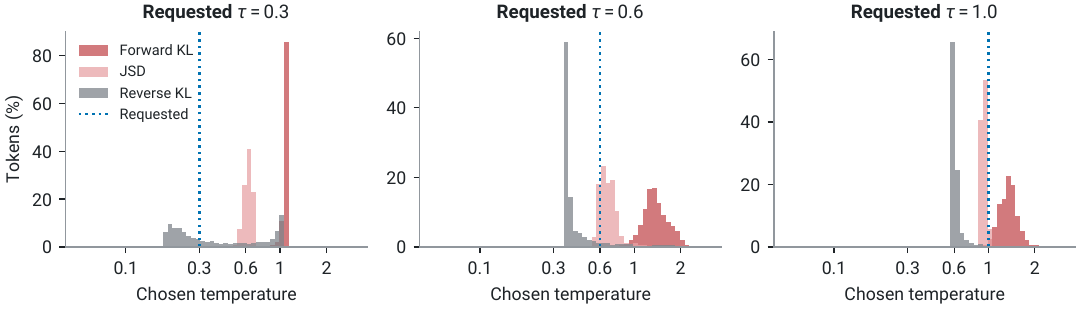}
\caption{Temperatures chosen by per-step TempQuant for GPTQ3 under each training loss, on 768 held-out steps.}
\label{fig:chosen-by-loss}
\end{figure}

Comparing the three also addresses a question this trade-off raises:
whether it is better to correct the flipped tokens, which are rare but carry
most of the error, or the preserved tokens, which carry less of it but are
far more common. Figures~\ref{fig:tradeoff} and~\ref{fig:chosen-by-loss} show the
temperatures each divergence selects for global and per-step TempQuant, and
Appendix~\ref{app:divergence-toy} gives the settings global TempQuant
selects at every requested temperature.

\subsection{Global TempQuant}
\label{sec:adaptive-results}

BF16 accuracy stays nearly flat across our three temperatures
(Table~\ref{tab:gptq3-main}), but under 3-bit quantization the deployed
temperature decides both accuracy and the divergence from BF16
(Figure~\ref{fig:tradeoff}). We first show that
the best temperature for the quantized model differs from the
full-precision one, as temperature tuning finds, and then that global
TempQuant, which does no sweep and only learns the setting that mimics BF16,
finds nearly the same temperature. Native 4-bit MXFP4 behaves similarly
(Table~\ref{tab:mxfp4-learned-global}).
Global TempQuant turns top-$p$ off at every requested temperature, which by
itself removes much of the forward KL, and the learned temperature removes
more (Appendix Table~\ref{tab:fitted-control-summary}).

\begin{figure}[!tb]
\centering
\includegraphics[width=\linewidth]{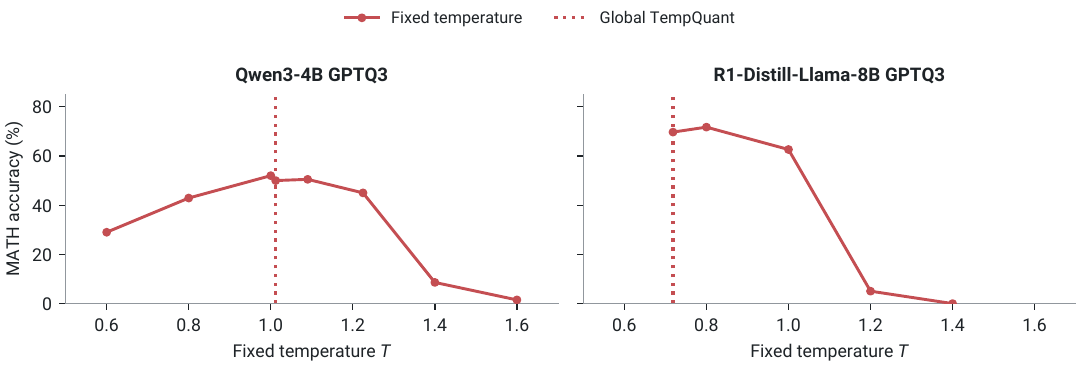}
\caption{MATH accuracy at fixed temperatures, and the temperature global TempQuant chooses from BF16 without answers for a requested $0.6$ (dotted).}
\label{fig:label-free}
\end{figure}

\paragraph{TempQuant versus temperature tuning.}
The usual way to choose a sampler setting is to tune it on labeled data:
generate and grade responses at every candidate setting and keep the most
accurate one \citep{shi2024decoding,wang2025autodeco}. Tuning GPTQ3 this way on 200 MATH questions took about 7,800 graded
responses and 170 million generated tokens (Figure~\ref{fig:label-free},
Appendix Figure~\ref{fig:label-free-3d}), and has to be repeated for every model and
task. Global TempQuant needs no labels or grading, only the full-precision
model's next-token probabilities on a few hundred unlabeled responses, and
lands close to the setting tuning finds. The best setting differs between
models, quantization methods, and requested temperatures, so a fixed rule
such as sampling every quantized model at a higher temperature does not carry over
(Appendix~\ref{app:fixed-one}).

\paragraph{TempQuant and the recommended recipe.}
Model providers have noticed that quantized models degrade and loop at these
temperatures. Qwen's cards for its quantized models keep the temperature at
$0.6$ but add a presence penalty of $1.5$ to stop endless repetition
\citep{qwen3card4bawq}, the failure that costs GPTQ3 most of its accuracy.
Global TempQuant performs better than this recipe, and adding the penalty on top
of global TempQuant helps a little further, so the two combine
(Table~\ref{tab:presence-penalty}). The penalty pushes the model away from
tokens it has already used, while TempQuant moves it toward the
full-precision model.

\begin{table}[!htbp]
\centering\small
\setlength{\tabcolsep}{4pt}
\begin{tabular*}{\linewidth}{@{\extracolsep{\fill}}lrrrrrr@{}}
\toprule
 & \multicolumn{3}{c}{MATH} & \multicolumn{3}{c}{GPQA}\\
\cmidrule(lr){2-4}\cmidrule(l){5-7}
Sampler & Accuracy & Strict & At limit & Accuracy & Strict & At limit\\
\midrule
Default $(0.6,.95)$ & 23.7 & 9.5 & 89.0 & 11.1 & 1.5 & 93.4\\
Qwen recipe, presence penalty & 33.8 & 26.0 & 67.0 & 17.7 & 11.1 & 65.7\\
\textbf{Global TempQuant} $(1.01,1)$ & \textbf{51.0} & \textbf{36.7} & \textbf{47.5} & \textbf{23.7} & \textbf{21.7} & \textbf{12.1} \\
\textbf{Global TempQuant} + presence penalty & \textbf{53.5} & \textbf{46.7} & \textbf{29.0} & \textbf{23.7} & \textbf{23.2} & \textbf{7.1} \\
\bottomrule
\end{tabular*}
\caption{Qwen3-4B GPTQ3 at $\tau=0.6$ (\%). Qwen's recipe adds top-$k$ 20 and presence penalty 1.5; strict counts responses at the length limit as wrong.}
\label{tab:presence-penalty}
\end{table}

\subsection{Per-step TempQuant}
\label{sec:per-step}
Per-step TempQuant chooses the setting for every token to avoid the
compromise of Section~\ref{sec:which-way}.
\label{sec:attainable-correction}To measure how much of the quantization
error this could remove, we test an oracle that knows BF16's distribution at
every token and picks the quantized model's temperature and top-$p$, from
the grid global TempQuant searches, to minimize the divergence. The oracle
removes most of the divergence quantization adds, 84\% of the forward KL at
the recommended $0.6$, almost all of it through the temperature alone
(Appendix Table~\ref{tab:fitted-control-summary}), and sampling GPTQ3 with
its settings brings GSM8K accuracy close to BF16's and lifts MATH well above
TempQuant (Appendix~\ref{app:downstream-oracle}). Per-step TempQuant tries
to capture this headroom from the quantized model alone, so it has to infer
from the quantized logits whether the top token changed (examples in
Appendix~\ref{app:per-step-examples}).

\subsection{Downstream accuracy}
\label{sec:downstream}
{\setlength{\intextsep}{4pt}\begin{table}[!htbp]
\centering\small
\setlength{\tabcolsep}{3.5pt}
\renewcommand{\arraystretch}{0.92}
\begin{tabular*}{\linewidth}{@{\extracolsep{\fill}}lrrrrrrr@{}}
\toprule
 & \multicolumn{4}{c}{Accuracy (\%)} & Mean & At length & \\
\cmidrule(lr){2-5}
Policy & GSM8K & MATH & AIME & GPQA & accuracy & limit (\%) & Closeness\\
\midrule
\multicolumn{8}{@{}l}{\textit{Temperature ($\tau$) = 0.3}}\\[1pt]
BF16 & 94.33 & 95.05 & 66.56 & 40.34 & 74.07 & 5.5 & 100\\
\cmidrule(lr){1-8}
GPTQ3 & 18.30 & 15.20 & 0.73 & 6.69 & 10.23 & 97.6 & 43.7\\
Global TempQuant & \gain{20}{\textbf{81.71}} & \gain{20}{\textbf{53.00}} & \textbf{1.98} & \gain{20}{25.00} & \textbf{40.42} & \textbf{43.5} & 71.0\\
Per-step TempQuant, forward KL & \gain{20}{81.33} & \gain{20}{52.10} & 1.43 & \gain{20}{\textbf{26.07}} & 40.23 & 44.4 & \textbf{71.2}\\
Per-step TempQuant, JSD & \gain{20}{52.72} & \gain{20}{29.45} & 0.49 & \gain{20}{11.05} & 23.43 & 83.1 & 49.9\\
Per-step TempQuant, reverse KL & 14.12 & 12.53 & 0.52 & 5.18 & 8.08 & 98.8 & 43.2\\
\midrule
\multicolumn{8}{@{}l}{\textit{Temperature ($\tau$) = 0.6}}\\[1pt]
BF16 & 94.83 & 95.75 & 69.58 & 41.48 & 75.41 & 2.6 & 100\\
\cmidrule(lr){1-8}
GPTQ3 & 50.36 & 29.90 & 0.94 & 10.98 & 23.05 & 83.7 & 47.3\\
Global TempQuant & \gain{20}{\textbf{80.80}} & \gain{20}{\textbf{52.35}} & \textbf{1.72} & \gain{20}{\textbf{25.19}} & \textbf{40.01} & \textbf{42.7} & \textbf{70.2}\\
Per-step TempQuant, forward KL & \gain{20}{79.68} & \gain{20}{51.35} & 1.43 & \gain{20}{24.08} & 39.14 & 48.2 & 68.1\\
Per-step TempQuant, JSD & \gain{20}{61.50} & \gain{20}{34.92} & 0.55 & 13.73 & 27.67 & 76.8 & 50.9\\
Per-step TempQuant, reverse KL & 20.29 & 16.88 & 0.57 & 6.60 & 11.08 & 96.9 & 41.7\\
\midrule
\multicolumn{8}{@{}l}{\textit{Temperature ($\tau$) = 1.0}}\\[1pt]
BF16 & 94.92 & 95.05 & 68.54 & 43.50 & 75.50 & 2.3 & 100\\
\cmidrule(lr){1-8}
GPTQ3 & 80.27 & 50.40 & 0.73 & 21.59 & 38.25 & 48.6 & 66.0\\
Global TempQuant & 78.66 & 49.00 & 1.77 & 22.16 & 37.90 & 44.3 & 70.1\\
Per-step TempQuant, forward KL & \textbf{80.86} & \textbf{51.90} & \textbf{2.29} & \textbf{25.28} & \textbf{40.08} & \textbf{42.2} & \textbf{72.1}\\
Per-step TempQuant, JSD & 79.25 & 47.50 & 0.65 & 21.53 & 37.23 & 53.6 & 63.0\\
Per-step TempQuant, reverse KL & 45.56 & 28.52 & 0.81 & 9.56 & 21.12 & 86.0 & 46.2\\
\bottomrule
\end{tabular*}
\caption{GPTQ3 accuracy, with the share of responses at the length limit and the closeness to BF16 averaged over the four tasks. Bold marks the best quantized value (lowest for the length limit); green marks Holm-significant gains over GPTQ3.}
\label{tab:gptq3-main}
\end{table}
}
Table~\ref{tab:gptq3-main} measures accuracy on each policy's own responses.
At $0.3$ and $0.6$, TempQuant raises GSM8K, MATH, and GPQA accuracy far
above raw GPTQ3, lifting mean accuracy from $10\%$ to $40\%$ at $0.3$. At
$1.0$, where raw GPTQ3 loops less, there is less to recover (ablations and
strict grading in Appendix~\ref{app:full-results}). The correction carries over to 3-bit
Qwen3-8B and to Gemma 4 31B, a different family, raising Gemma's MATH
accuracy from 70\% to 90\%, but not to R1-Distill-Llama-8B, which rarely
runs into repetition (Appendix Table~\ref{tab:cross-model}).
Table~\ref{tab:gptq3-main} also reports how close each policy's generated
text is to BF16's with a closeness score: 100 minus the average gap between
a policy and BF16 on four statistics of its text, namely word overlap with
BF16's responses, how alike the policy's own responses to the same prompt
are, repeated four-grams, and the share of responses that hit the length
limit (details in Appendix~\ref{app:gptq3-full-downstream}).

Of the three divergences, forward KL gives per-step TempQuant the highest
accuracy, matching global TempQuant at $0.3$ and $0.6$ and exceeding it at
$1.0$ (Appendix Table~\ref{tab:gptq3-global-significance}). Global TempQuant
adds no cost at deployment, and per-step TempQuant adds time per token
(Appendix~\ref{app:runtime-timing}).

\subsection{Does TempQuant preserve behavior and task performance?}
\label{sec:behavior-results}
Since similar accuracy can hide large changes in behavior
\citep{dutta2024accuracy}, and matching a reference model need not improve
task performance \citep{stanton2021distillation}, we report behavior
alongside accuracy, using task-specific measures and the closeness score
of Section~\ref{sec:downstream}. The three divergences behave differently here, and the results separate into
two cases: long reasoning under heavy degradation, and shorter outputs or
lighter degradation.

\subsubsection{Long reasoning under heavy degradation}
At the requested temperature, GPTQ3's reasoning collapses into repetition:
most of its responses repeat themselves until the length limit, while BF16's
rarely do (Table~\ref{tab:gptq3-main}). Here, preserving BF16's
behavior mostly means stopping these loops, and a response that never
finishes cannot be graded correct. Breaking the loops and giving flipped
tokens back their probability both need a higher temperature. Forward KL,
which warms the most, undoes about half of the collapse and recovers the
most accuracy, while reverse KL cools, keeps the model on its favorite
tokens, and makes the repetition worse (Table~\ref{tab:gptq3-main},
Appendix~\ref{app:gptq3-full-downstream}). The same pattern holds on the
other 3-bit models (Appendix Table~\ref{tab:cross-model}). In this case the forward KL
controller preserves BF16's behavior best, because it targets the largest
source of divergence, the looping. Quantized reasoning models also
overthink \citep{lotfi2026overthinking}, and TempQuant reduces this only in
part (Appendix~\ref{app:overthinking}).

\subsubsection{Short text and light degradation}
On short summaries, GPTQ3 does not loop, so there is little accuracy to
recover. Its main error is instead spreading probability onto tokens BF16
would not pick, so at $0.3$ it varies about as much as BF16 does at $0.6$.
Reverse KL penalizes exactly this error, so it cools GPTQ3 back toward
BF16's variety, while forward KL makes nearly every sample different. The
reverse KL controller is closest to BF16 on variety, meaning diversity,
summary quality, and closeness at every temperature
(Table~\ref{tab:xsum-diversity}).

\begin{figure}[!htb]
\centering
\includegraphics[width=\linewidth]{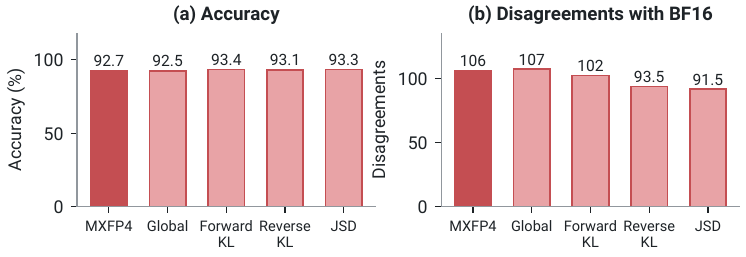}
\caption{MXFP4 on MATH at $\tau=0.6$: (a) accuracy and (b) disagreements with BF16, samples that one answers correctly and the other does not.}
\label{fig:correctness-flips}
\end{figure}

MXFP4 is a similar case on reasoning, with lighter degradation. It loses
little accuracy, but similar accuracy can hide many answer flips, samples that the full-precision
model answers correctly and the quantized model does not, or the reverse
\citep{dutta2024accuracy}.
On MATH at $0.6$, the reverse KL and JSD controllers keep accuracy about the
same while removing about a third of the extra answer flips MXFP4 has with BF16
(Figure~\ref{fig:correctness-flips}), and across all four tasks and three
temperatures, the JSD controller brings which questions MXFP4 answers
correctly closer to BF16's (Appendix Table~\ref{tab:mxfp4-pooled-test}).

These results suggest the forward KL controller for more aggressive
quantization, the JSD controller for reasoning under light degradation, and
the reverse KL controller for short text, though larger experiments are
needed before committing to this split.

\begin{table}[!htbp]
\centering\small
\setlength{\tabcolsep}{6pt}
\renewcommand{\arraystretch}{0.95}
\begin{tabular*}{\linewidth}{@{\extracolsep{\fill}}lrrrr@{}}
\toprule
Policy & \shortstack{Unique\\summaries (\%)} & \shortstack{Meaning\\diversity} & ROUGE-2 & Closeness\\
\midrule
\multicolumn{5}{@{}l}{\textit{Temperature ($\tau$) = 0.3}}\\[1pt]
BF16 (value) & 50 & 0.037 & 6.81 & 100\\
\cmidrule(lr){1-5}
$\Delta$ GPTQ3 & +22 & +0.045 & $-$0.78 & $-$15.2\\
$\Delta$ Per-step TempQuant, forward KL & +49 & +0.141 & $-$1.22 & $-$21.3\\
$\Delta$ Per-step TempQuant, JSD & +43 & +0.090 & $-$0.99 & $-$17.8\\
$\Delta$ Per-step TempQuant, reverse KL & \textbf{+13} & \textbf{+0.032} & \textbf{$-$0.71} & \textbf{$-$14.8}\\
\midrule
\multicolumn{5}{@{}l}{\textit{Temperature ($\tau$) = 0.6}}\\[1pt]
BF16 (value) & 75 & 0.061 & 6.52 & 100\\
\cmidrule(lr){1-5}
$\Delta$ GPTQ3 & +17 & +0.067 & $-$0.61 & $-$13.6\\
$\Delta$ Per-step TempQuant, forward KL & +25 & +0.163 & $-$1.24 & $-$21.2\\
$\Delta$ Per-step TempQuant, JSD & +23 & +0.089 & $-$0.74 & $-$15.9\\
$\Delta$ Per-step TempQuant, reverse KL & \textbf{+1} & \textbf{+0.028} & \textbf{$-$0.46} & \textbf{$-$11.7}\\
\midrule
\multicolumn{5}{@{}l}{\textit{Temperature ($\tau$) = 1.0}}\\[1pt]
BF16 (value) & 90 & 0.092 & 6.25 & 100\\
\cmidrule(lr){1-5}
$\Delta$ GPTQ3 & +8 & +0.074 & $-$0.53 & $-$12.2\\
$\Delta$ Per-step TempQuant, forward KL & +10 & +0.121 & $-$0.89 & $-$16.5\\
$\Delta$ Per-step TempQuant, JSD & +8 & +0.071 & $-$0.55 & $-$12.1\\
$\Delta$ Per-step TempQuant, reverse KL & \textbf{0} & \textbf{+0.029} & \textbf{$-$0.36} & \textbf{$-$9.3}\\
\bottomrule
\end{tabular*}
\caption{GPTQ3 XSum summaries, eight per article: share of unique summaries, meaning diversity (mean embedding distance between samples), quality (ROUGE-2 against human summaries), and closeness to BF16 (Section~\ref{sec:downstream}). BF16 rows give its values, and $\Delta$ rows give each policy's change from BF16, so values nearer 0 are closer. Bold marks the closest policy.}
\label{tab:xsum-diversity}
\end{table}

% === Limitations ===
\section{Limitations}
\label{sec:limitations}
Temperature and top-$p$ cannot reorder tokens, so errors that flip rankings
remain even for the oracle. The accuracy gains are large only when the quantized model runs into
repetition. None of the per-step controllers we trained preserves every
behavior we measure. We fit TempQuant on math text only, evaluate reasoning
with thinking mode on, use a single reference top-$p$ of $.95$, and compare
with Qwen's presence penalty on one model. Global TempQuant adds no cost at
deployment, while our unoptimized per-step implementation adds 18--34\% to
the time per generated token (Appendix~\ref{app:runtime-timing}).

% === Conclusion ===
\section{Conclusion}
Quantized models run hotter, but most of the error sits on the few tokens
whose ranking quantization flips, which tend to need warming while the rest
tend to need cooling. Global TempQuant finds without labels the setting that tuning on
answers would find, and recovers much of the accuracy lost by 3-bit models
that fall into repetition. Per-step TempQuant, which chooses the setting at
every token from the quantized model alone, matches this at requested
temperatures $0.3$ and $0.6$ and is more accurate at $1.0$. The choice of divergence decides what is preserved:
forward KL restores accuracy when the model loops, while for a lightly
degraded model reverse KL brings short text and JSD brings reasoning closer
to BF16's. Overall, we show
that much of the quantization error can be recovered through temperature
and top-$p$, so quantized models should be deployed at different settings
from their full-precision counterparts.

\begingroup
\setlength{\bibsep}{2pt}
\bibliography{references_main,references_working}
\bibliographystyle{mlsys2025}
\endgroup
\clearpage
\appendix

\section{Extended related work}
\label{app:related}

\paragraph{Adaptive decoding.}
Adaptive temperature scaling (ATS) and TempNet also learn temperatures at a
fine granularity, but for different reasons. ATS learns a temperature for
each token to improve the model's confidence calibration
\citep{xie2024ats}, and TempNet learns to predict a temperature for each
input to improve prediction and generalization across tasks
\citep{qiu2024tempnet}. ReSET also sets the temperature per token, and
targets NVFP4-quantized reasoning models. It observes that quantization adds
sampling errors at confident tokens and concentrates probability too much
at uncertain reasoning steps. It samples high-entropy tokens at temperature
$1.0$ and the rest at a lower temperature, with an entropy threshold that
adapts to the current reasoning step, and calibrates the low temperature for
each model and task on labeled held-out data \citep{lee2026reset}. It does
not learn a temperature predictor, and its goal is to balance precision and
exploration during reasoning, rather than to imitate the full-precision
model. Our entropy-threshold baseline in Appendix~\ref{app:entropy-rules}
is a simpler rule in the spirit of ReSET, and not ReSET itself. AutoDeco is
the closest work to ours architecturally, since it also learns temperature
and top-$p$ at each generation step, with two small heads on the language
model. It trains them with cross-entropy to raise the probability of the
next token in reasoning traces selected for correct final answers
\citep{wang2025autodeco}, and we borrow its two-head design. TempQuant
instead learns these settings so that the quantized model's next-token
probabilities match those of its full-precision counterpart, and uses them
to measure how much of the quantization error the sampler can remove.

\paragraph{Decoding-time fixes for degeneration.}
\citet{lotfi2026overthinking} measure the token-level KL between quantized
and full-precision reasoning models, find that it is high at uncertain
positions, where quantized models oversample overthinking markers such as
``wait'', and reduce this with a training-free logit penalty on a curated
set of these markers. TempQuant also corrects the quantized model at
decoding time without retraining it, but fits the standard temperature and
top-$p$ to the full-precision model, with no chosen set of tokens. More
general samplers target repetition and incoherence without a reference
model: repetition penalties \citep{keskar2019ctrl}, Mirostat, which holds
perplexity at a target \citep{basu2021mirostat}, locally typical sampling
\citep{meister2023typical}, truncation sampling
\citep{hewitt2022truncation}, and min-$p$ \citep{nguyen2025minp}.
Repetition also reinforces itself once it starts \citep{xu2022loop}, which
is consistent with how quickly a cold 3-bit model falls into loops.
\citet{braverman2020calibration} show that the entropy rate of generated
text drifts away from that of real text, and correct this with a single
temperature-like parameter, a close precedent for global TempQuant. We fit
such a parameter against the full-precision model after quantization.

\paragraph{Temperature calibration.}
Other work learns temperatures to calibrate a model's confidence. Temperature
scaling fits one temperature on labeled data \citep{guo2017calibration}.
DACA fits a temperature with a reference-based forward KL so that a
post-trained model's confidence matches its pre-trained model
\citep{luo2025daca}. Thermometer predicts one temperature for a
task by aggregating unlabeled prompts \citep{shen2024thermometer}, and
Dual-Align combines output and layerwise supervision to fit a single
temperature \citep{luo2026dualalign}. We do not select temperature and
top-$p$ for confidence calibration, but to preserve the behavior of a
quantized model.

\paragraph{Distillation.}
Matching a teacher's next-token distribution is the basis of knowledge
distillation \citep{hinton2015distilling}. LLM-QAT applies it to
quantization, updating the quantized weights to minimize forward KL to the
full-precision model \citep{liu2023llmqat}, and \citet{agarwal2024gkd}
study how the choice of divergence affects a student that cannot reproduce
its teacher.

\paragraph{Temperature and generation behavior.}
Temperature affects the diversity and coherence of generated text.
\citet{holtzman2020degeneration} show that low-temperature sampling can
reduce diversity and increase repetitive loops in open-ended generation. In
a Llama-2 story-generation study, \citet{peeperkorn2024temperature} find
that higher temperature is only weakly associated with novelty and more
strongly associated with incoherence. For full-precision models, the
temperature between $0$ and $1$ has little effect on problem-solving
accuracy \citep{renze2024temperature}.

\paragraph{Behavioral effects of quantization.}
Quantization can change reasoning behavior as well as accuracy, and
similar aggregate accuracy does not establish that behavior is preserved
\citep{dutta2024accuracy}. \citet{lotfi2026overthinking} find that
aggressive quantization can lengthen chains of thought and increase
overthinking errors, where a model reaches a correct intermediate answer
and then abandons it. In dialogue evaluations,
\citet{li2024evaluatingquantized} document sentence-level repetition under
3-bit weight or KV-cache quantization and repeated-token failures at 2
bits. The effects vary by configuration: \citet{liu2025quantizationhurts}
find response lengths close to BF16 under several settings, while more
aggressive quantization can lengthen outputs. In our data, 3-bit responses
are several times longer than full-precision ones, mostly because they run
into repetition. We measure response length and repetition alongside
accuracy, using BF16 as our behavioral reference.

\paragraph{Quantization and token probabilities.}
\citet{wu2026decisions} find that low-bit quantization can shrink decision
margins, the logit differences between the first tokens of two candidate
continuations, and introduce directional biases that reverse model
preferences. A margin describes a particular choice, while entropy
describes uncertainty over the full vocabulary, and one does not determine
the other. We study the related question of how quantization changes
entropy and which tokens' rankings it flips. Our quantized models use GPTQ
\citep{frantar2023gptq}, AWQ \citep{lin2024awq}, and the MXFP4 microscaling
format \citep{rouhani2023microscaling}.

\section{Why the divergences disagree on flipped tokens}
\label{app:divergence-toy}
\begin{figure}[!htbp]
\centering
\includegraphics[width=.88\linewidth]{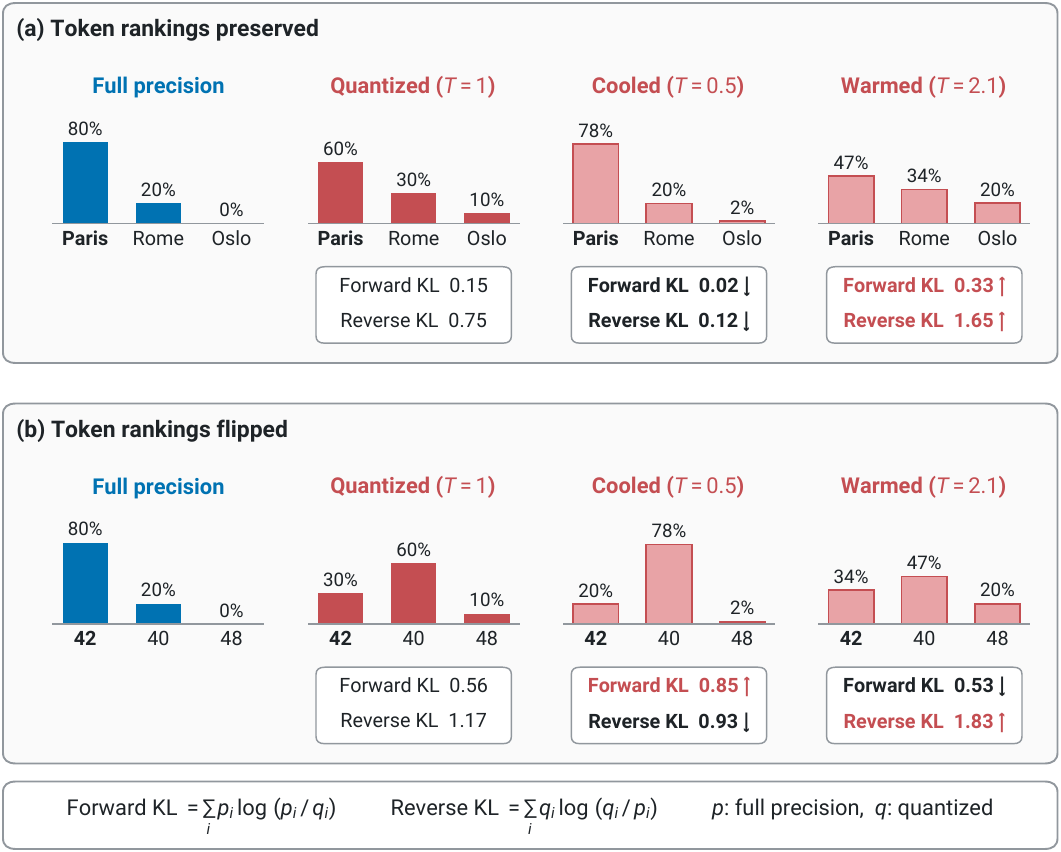}
\caption{Forward and reverse KL for the three-token examples of Figure~\ref{fig:temperature-mechanism}, before correction and after cooling or warming the quantized model. Arrows mark whether each divergence falls or rises.}
\label{fig:divergence-toy}
\end{figure}

Figure~\ref{fig:divergence-toy} uses the three-token examples of
Figure~\ref{fig:temperature-mechanism} to show how cooling and warming the
quantized model change forward KL, $\sum_i p_i\log(p_i/q_i)$, and reverse KL,
$\sum_i q_i\log(q_i/p_i)$, where $p$ is the full-precision distribution and
$q$ the quantized one. When the rankings are preserved, quantization only
spreads the probabilities, and cooling brings $q$ back toward $p$ under both
divergences. When the rankings flip, cooling sharpens $q$ around the wrong
token: reverse KL improves, because $q$ moves its mass off the token BF16
rules out, while forward KL worsens, because BF16's preferred token loses
probability. Warming does the opposite. A typical preserved token therefore pulls
a single setting the same way under both divergences, and the divergences
part ways mainly on the flipped tokens.

Table~\ref{tab:learned-global-objectives} gives the settings global
TempQuant selects under each divergence. Unlike
Figure~\ref{fig:tradeoff}, which varies the temperature alone, the search
also chooses top-$p$, which lets reverse KL cut more tokens and cool
further.

\begin{table}[!htbp]
\centering\small
\setlength{\tabcolsep}{6pt}
\begin{tabular*}{\linewidth}{@{\extracolsep{\fill}}lrrr@{}}
\toprule
Training divergence & $\tau=0.3$ & $\tau=0.6$ & $\tau=1.0$\\
\midrule
Forward KL & $(1.09,1.00)$ & $(1.13,1.00)$ & $(1.22,1.00)$\\
JSD & $(0.58,0.98)$ & $(0.67,0.99)$ & $(0.82,0.98)$\\
Reverse KL & $(0.20,0.90)^\ast$ & $(0.15,0.90)^\ast$ & $(0.15,0.90)^\ast$\\
\bottomrule
\end{tabular*}
\caption{Temperature and top-$p$ $(T,p)$ that global TempQuant selects for GPTQ3 under each training divergence. $^\ast$At the lower edge of the search range.}
\label{tab:learned-global-objectives}
\end{table}

\section{Controller architecture and training}
\label{app:final-spec}
\begin{figure}[!htb]
\centering
\includegraphics[width=.75\linewidth]{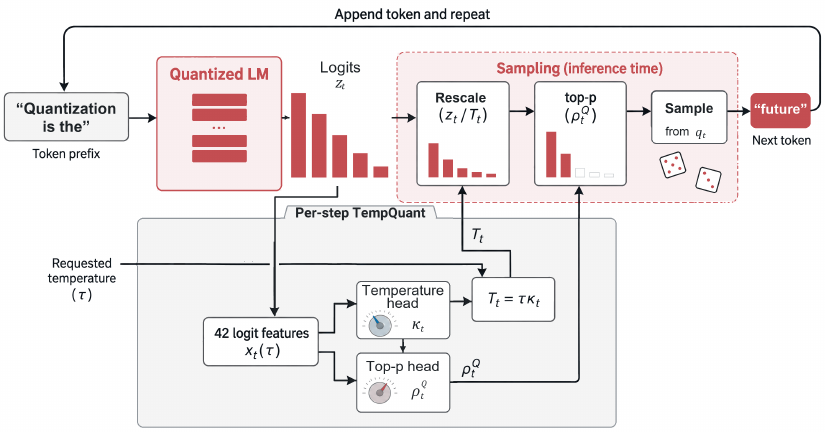}
\caption{Per-step TempQuant's two prediction heads and their connection to next-token sampling.}
\label{fig:inference}
\end{figure}

We train two controllers per objective with different seeds and average
their results. Figure~\ref{fig:inference} shows the design.

\subsection{Architecture and inputs}
The temperature and top-$p$ heads share the same 42 logit features but have
separate weights. The temperature head outputs the correction factor
$\kappa$, and the top-$p$ head also receives $\log\kappa$, so gradients reach
both heads. We pass $\log\kappa$ rather than $\kappa$ so that keeping the
requested temperature gives zero, and doubling or halving it moves the input
by the same amount in opposite directions. Following AutoDeco, the top-$p$
head receives the predicted temperature because changing the temperature
changes which tokens a given cutoff keeps.

\paragraph{Model and quantization.}
We use
\href{https://huggingface.co/Qwen/Qwen3-4B/tree/1cfa9a7208912126459214e8b04321603b3df60c}{\texttt{Qwen/Qwen3-4B}}
\citep{qwen3technicalreport}. MXFP4 quantizes the linear weights statically
and their inputs dynamically, in groups of 32 with shared E8M0 scales.
Embeddings, normalization, the tied vocabulary head, and the KV cache stay
in BF16. We serve models with vLLM \citep{kwon2023pagedattention}.

\begin{table}[!htbp]
\centering
\renewcommand{\arraystretch}{1.1}
\begin{tabularx}{\linewidth}{@{}lX@{}}
\toprule
\multicolumn{2}{@{}l@{}}{\textbf{Controller architecture}}\\
\midrule
Model & Qwen3-4B GPTQ3 (main results) or native MXFP4 W4A4; BF16 KV \\
Reference & BF16 temperatures $0.3,0.6,1.0$, top-$p=.95$\\
Input & Ten distribution summaries followed by 32 ranked logit features\\
Temperature head & $42\to1398\to16\to1$, Tanh hidden activations\\
Top-$p$ head & $43\to1375\to16\to1$, Tanh\\
Trainable parameters & 165,048\\
Temperature output & $\kappa\in(0.083,83)$\\
Top-$p$ output & $\rho^Q\in(0,1)$\\
Sampling temperature & $T=\tau\kappa$, the requested temperature $\tau$
multiplied by the correction factor\\
\bottomrule
\end{tabularx}
\caption{Architecture and output ranges of the per-step predictor.}
\label{tab:final-spec}
\end{table}

\subsection{Input features}
Table~\ref{tab:logit-features} lists the features. We adjust the logits for
the requested temperature first, and normalize each feature with its
training mean and standard deviation. Here $\widetilde z$ is the logits
scaled by $\tau_0/\tau$ and shifted so the maximum is zero, with
$\tau_0=0.6$, $r_k$ is its $k$th largest value, $s$ is its standard
deviation, and $v=\softmax(\widetilde z/0.6)$.

\begin{table}[!htbp]
\centering
\renewcommand{\arraystretch}{1.15}
\begin{tabularx}{\linewidth}{@{}rXl@{}}
\toprule
Coordinate & Feature & Definition\\
\midrule
1 & Gap between the first- and second-ranked logits & $r_1-r_2$\\
2 & Gap between the first- and eighth-ranked logits & $r_1-r_8$\\
3 & Gap between the first- and 32nd-ranked logits & $r_1-r_{32}$\\
4 & Logit sample standard deviation & $\operatorname{std}(\widetilde z)$\\
5 & Mean logit & $V^{-1}\sum_i\widetilde z_i$\\
6 & Maximum logit after centering & $r_1=0$\\
7 & Entropy (nats) & $H(v)=-\sum_i v_i\log v_i$\\
8 & Largest token probability & $v_{(1)}$\\
9 & Probability mass of the eight most likely tokens & $\textstyle\sum\nolimits_{k=1}^{8}v_{(k)}$\\[2pt]
10 & Probability mass of the 32 most likely tokens & $\textstyle\sum\nolimits_{k=1}^{32}v_{(k)}$\\[2pt]
\midrule
11--42 & The 32 largest logits divided by their scale & $r_k/s,\quad k=1,\ldots,32$\\
\bottomrule
\end{tabularx}
\caption{The controller's ten distribution summaries and 32 leading-logit features, before normalization.}
\label{tab:logit-features}
\end{table}

\subsection{Output ranges and starting values}
Each head produces an unbounded number, and a sigmoid maps it into an
allowed range. For head outputs $a_\theta(x)$ and $b_\phi(x,\log\kappa)$,
\begin{align}
 \log\kappa&=\log\kappa_{\min}
 +\sigma(a_\theta(x)+c_T)\log\!\left(\frac{\kappa_{\max}}{\kappa_{\min}}\right),
 \nonumber\\
 \rho^Q&=\sigma(b_\phi(x,\log\kappa)+c_p),
 \qquad (\kappa_{\min},\kappa_{\max})=\left(\tfrac{0.05}{0.6},\tfrac{50}{0.6}\right).
 \label{eq:final-parameterization}
\end{align}
The first line places $\log\kappa$ between $\log\kappa_{\min}$ and
$\log\kappa_{\max}$, so the correction lies between about $0.083$ and $83$
and is spread evenly on a log scale. Equivalently, the temperature the head
would use for a request of $0.6$, which is the scale the features are
normalized to, lies between $0.05$ and $50$. A correction of $\kappa=1$
keeps the requested temperature, and the sampler uses $T_t=\tau\kappa_t$.
The second line keeps the top-$p$ cutoff between $0$ and $1$. These are
$F_\theta$ and $G_\phi$ in Equation~\ref{eq:final-heads}.

The constants $c_T$ and $c_p$ are set so that, before training, both heads
output the global TempQuant setting fitted on the training data, and the
final layers start at zero. Every prefix therefore starts at that global
setting, and training learns how far to move away from it at each token.

\subsection{Training settings}
Only the two heads are trained (Table~\ref{tab:training-spec}). We use the
controller after 120 updates.

\begin{table}[!htbp]
\centering
\renewcommand{\arraystretch}{1.1}
\begin{tabularx}{\linewidth}{@{}lX@{}}
\toprule
\multicolumn{2}{@{}l@{}}{\textbf{Training configuration}}\\
\midrule
Objectives & Smoothed forward KL, smoothed reverse KL, symmetric JSD\\
Loss scale $c_D$ & $1/4$ for either KL objective; $1$ for JSD\\
Uniform mixture $\eta$ & $10^{-4}$ for KL; none for JSD\\
Smooth top-$p$ width & $\epsilon=.01$\\
Action-variance weight & $\lambda=.01$\\
Optimizer & Adam: $\beta_1=.9$, $\beta_2=.999$,
$\epsilon_{\rm Adam}=10^{-8}$, weight decay $0$;
without AMSGrad\\
Learning rate & $.001$ for updates 1--30; $.0001$ for 31--120;
optimizer state retained at the transition\\
Update / clipping & Full weighted pool of 6,912 rows, accumulated in
chunks of 32; global gradient norm clipped to 1 before each update\\
\bottomrule
\end{tabularx}
\caption{Training settings for all three objectives and every training run.}
\label{tab:training-spec}
\end{table}

\paragraph{Action-variance regularization.}
A small penalty keeps the predicted settings from varying too much from one
training prefix to the next. At each requested temperature, it adds the
variance across prefixes of the log predicted temperature and of the
predicted cutoff on the log-odds scale, which spreads out cutoffs near $1$,
and then averages over the three requested temperatures:
\begin{equation}
 \mathcal R=\frac13\sum_{\tau\in\{0.3,0.6,1.0\}}\frac1N
 \sum_{t=1}^{N}\left[(\ell_{t,\tau}-\bar\ell_\tau)^2
 +(\zeta_{t,\tau}-\bar\zeta_\tau)^2\right],
 \label{eq:action-variance}
\end{equation}
where $\ell_{t,\tau}=\log\kappa_t(\tau)$, $\zeta_{t,\tau}=\operatorname{logit}\rho_t^Q(\tau)$,
$N=2{,}304$ training prefixes are weighted equally, and bars are means. The
implementation scales the divergence by $c_D$ and the penalty by
$\lambda=.01$, which gives Equation~\ref{eq:final-objective} with
$\lambda_D=\lambda/c_D$.

\paragraph{Smooth top-$p$.}
During training we replace top-$p$ with a smooth version. Sort the logits in
ascending order, let $b_i$ be their probabilities after temperature
scaling, and let $C_i=\sum_{j\leq i}b_j$. With $\epsilon=0.01$,
\begin{equation}
 \log g_i=\log\sigma\!\left(\frac{C_i-(1-\rho^Q)}{\epsilon}\right)
          -\log\sigma\!\left(\frac{C_i}{\epsilon}\right),\qquad
 \widetilde q_i=\frac{b_i g_i}{\sum_j b_jg_j},
 \label{eq:smooth-nucleus}
\end{equation}
with $g_i=1$ for the top token. Targets, evaluation, and generation use
ordinary top-$p$. The KL smoothing weight and the sharpness of this smooth
top-$p$ are choices we make only so that training works, so
Table~\ref{tab:sensitivity} tests how much the results depend on them.
Halving or doubling either one changes held-out forward KL by at most $5\%$ and the
mean chosen temperature by at most $4\%$. The predictor therefore learns
essentially the same settings whatever values we pick, so these choices
make training possible without shaping what it learns. Since the held-out
divergence is measured with ordinary top-$p$, this also shows that training
with the smooth version carries over to the hard cutoff used in deployment.

\begin{table}[!htbp]
\centering\small
\setlength{\tabcolsep}{4pt}
\begin{tabular*}{\linewidth}{@{\extracolsep{\fill}}lrrrrrr@{}}
\toprule
 & \multicolumn{3}{c}{Forward KL} & \multicolumn{3}{c}{Mean chosen $T$}\\
\cmidrule(lr){2-4}\cmidrule(l){5-7}
Training variant & $\tau=0.3$ & $0.6$ & $1.0$ & $\tau=0.3$ & $0.6$ & $1.0$\\
\midrule
As in the paper & 0.594 & 0.483 & 0.482 & 1.090 & 1.404 & 1.371\\
KL smoothing $\times 0.5$ & 0.594 & 0.482 & 0.487 & 1.096 & 1.407 & 1.344\\
KL smoothing $\times 2$ & 0.594 & 0.482 & 0.488 & 1.097 & 1.407 & 1.345\\
Smooth top-$p$ sharpness $\times 0.5$ & 0.595 & 0.495 & 0.505 & 1.080 & 1.373 & 1.316\\
Smooth top-$p$ sharpness $\times 2$ & 0.595 & 0.467 & 0.468 & 1.105 & 1.428 & 1.344\\
\bottomrule
\end{tabular*}
\caption{GPTQ3 held-out forward KL and mean chosen temperature for forward-KL per-step TempQuant when the KL smoothing weight or the sharpness of the smooth top-$p$ used in training is halved or doubled. All variants use the same random seed; the chosen top-$p$ stays between $.974$ and $.976$.}
\label{tab:sensitivity}
\end{table}

\subsection{Training data}
\label{app:training-data}
We train on 24 GSM8K \citep{cobbe2021verifiers} and 24 MATH
\citep{hendrycks2021math} training questions, and hold out 32 more from
each, excluding MATH-500. For each question we collect two responses each
from BF16, uncorrected quantization, and a preliminary version of per-step
TempQuant, at
temperature $0.6$ and top-$p$ $0.95$. We take eight points from each training
response and two from each held-out response, giving 2,304 training steps
and 768 held-out steps, and reuse them at all three requested temperatures.

\section{Evaluation details}
\label{app:campaign-protocol}
\label{app:final-evaluation}

\subsection{Protocol notes for Section~\ref{sec:final-evaluation}}
\label{app:protocol-notes}
Fitting global TempQuant has a one-time cost: generating the calibration
responses and recording both models' next-token probabilities on them. Each
requested temperature needs its own fit, on the same data.

Every learned setting is fit on training questions, and divergence is
measured on 768 decoding steps from 64 other questions. Since we also used
these questions during development, we confirm the reductions on 64 fresh
questions with the same settings and no refitting: at $0.6$, GPTQ3's forward
KL falls from $1.35$ to $0.35$ with global TempQuant and $0.37$ with per-step
TempQuant, and MXFP4's from $0.79$ to $0.23$ and $0.32$. Unless noted, downstream results
for per-step TempQuant average two training runs, and its divergence results
on Qwen3-4B average five. The comparison with Qwen's recipe
(Table~\ref{tab:presence-penalty}) and the temperature sweep
(Figure~\ref{fig:label-free}) use 200 MATH questions, and the former also
198 GPQA questions, with one sample each.

All GPTQ3 models use 3-bit GPTQ with group size 128 and keep the vocabulary
head in BF16. The Qwen3-4B GPTQ3 checkpoint is a public third-party
release\footnote{\url{https://huggingface.co/irish-quant/Qwen-Qwen3-4B-3bit}, revision \texttt{d2c2902}.},
and we quantized Qwen3-8B, R1-Distill-Llama-8B, and Gemma 4 31B ourselves,
calibrating on C4 \citep{raffel2020c4}. GPTQ3 weights are dequantized for
BF16 computation, while MXFP4 runs in native four-bit computation.

\subsection{Benchmarks and comparison methods}
Table~\ref{tab:benchmark-coverage} lists the tasks. GSM8K, MATH-500, and
GPQA-Diamond \citep{rein2023gpqa} use their full test sets. AIME-120 is AIME90
plus the 30 problems from 2025. XSum \citep{narayan-etal-2018-dont} uses a
fixed validation subset screened for overlap with the training data. The
global TempQuant settings are searched per requested temperature
(Appendix~\ref{app:learned-global}), and per-step TempQuant ($T$ only)
keeps $p=.95$.

\begin{table}[!htbp]
\centering\small
\begin{tabular*}{\linewidth}{@{\extracolsep{\fill}}lrrr@{}}
\toprule
Task & Items & Trials & Output limit (tokens)\\
\midrule
GSM8K test & 1,319 & 4 & 16,384\\
MATH-500 test & 500 & 4 & 32,768\\
AIME-120 & 120 & 8 & 32,768\\
GPQA-Diamond & 198 & 8 & 32,768\\
XSum validation subset & 128 & 8 & 512\\
\bottomrule
\end{tabular*}
\caption{Evaluation tasks, responses per item, and output limits at each requested temperature.}
\label{tab:benchmark-coverage}
\label{tab:generation-settings}
\end{table}

\subsection{Generation and scoring}
\label{app:generation-scoring}
Seeds match across policies by task, question, and trial. Accuracy is the
average over sampled responses. Thinking mode is on for reasoning tasks and
off for XSum, with no repetition penalty. GSM8K and AIME use the Qwen2.5-Math
scorer \citep{yang2024qwenmath}, MATH-500 uses Math-Verify on the last boxed
answer, and GPQA uses LM-eval's flexible extraction. XSum quality is ROUGE-2
F1 against the human summaries \citep{lin-2004-rouge}, after removing a
leading thinking block. The MXFP4 and GPTQ3 studies each sample their own BF16 reference runs, so BF16 scores differ slightly between their tables.

\subsection{Behavior measures and uncertainty}
\label{app:behavior-protocol}
Length counts generated tokens, and the cap rate is the share of responses
that hit the length limit. BLEU against BF16
\citep{papineni-etal-2002-bleu,post-2018-call} measures word overlap with
BF16 responses, splitting each question's responses into two halves, so
BF16 against itself scores below 100. Self-BLEU measures how similar a
model's responses to the same question are, so higher means less diverse.
Repetition is the share of repeated four-grams within a response.

\paragraph{Testing one comparison.}
To compare two policies, for example raw GPTQ3 and global TempQuant, we use
the fact that both answer the same questions with the same random seeds, and
take the difference in accuracy question by question. We then resample the
questions with replacement many times and recompute the average difference
each time. The middle 95\% of these averages gives the interval, and a gain
is significant when the interval excludes zero. We call this a paired
bootstrap over questions. Intervals in the appendix are not adjusted for
multiple comparisons unless stated. When a grade is unknown, we count it as
half correct, the midpoint of counting it wrong or right.

\paragraph{Testing many comparisons.}
\label{app:multiplicity}
We compare several policies, each training loss and global TempQuant, on
several tasks and temperatures, and with this many comparisons a few would
look significant by chance. The Holm correction \citep{holm1979simple}
guards against this. It sorts the comparisons from smallest to largest
$p$-value and tests them in that order: with $m$ comparisons, the smallest
must fall below $0.05/m$, the next below $0.05/(m-1)$, and so on, and
testing stops at the first comparison that fails. The chance that any
comparison is declared significant by luck then stays below $5\%$ across
all of them together. For GPTQ3, we apply it across all
75 comparisons of every task, temperature, and policy against raw GPTQ3 and
against each other (Table~\ref{tab:gptq3-global-significance}). For MXFP4,
we apply it across its 54 comparisons of every task, temperature, and
training loss against native MXFP4 (Table~\ref{tab:objective-benchmarks}).
MATH drops 3 questions whose reference answers cannot be graded.

\begin{table}[!htbp]
\centering\small
\setlength{\tabcolsep}{3pt}
\begin{tabular*}{\linewidth}{@{\extracolsep{\fill}}llrrr@{}}
\toprule
Task & $\tau$ & Global $-$ GPTQ3 & Per-step $-$ GPTQ3 & Per-step $-$ global\\
\midrule
GSM8K & 0.3 & \gain{20}{$+63.5$ {\scriptsize[+61.7, +65.2]}} & \gain{20}{$+63.1$ {\scriptsize[+61.4, +64.8]}} & $-0.4$ {\scriptsize[-1.3, +0.5]}\\
 & 0.6 & \gain{20}{$+30.5$ {\scriptsize[+28.6, +32.3]}} & \gain{20}{$+29.3$ {\scriptsize[+27.5, +31.2]}} & $-1.1$ {\scriptsize[-2.1, -0.1]}\\
 & 1.0 & $-1.6$ {\scriptsize[-2.9, -0.4]} & $+0.6$ {\scriptsize[-0.4, +1.6]} & \gain{20}{$+2.2$ {\scriptsize[+1.1, +3.3]}}\\
\addlinespace
MATH & 0.3 & \gain{20}{$+38.0$ {\scriptsize[+34.5, +41.6]}} & \gain{20}{$+37.1$ {\scriptsize[+33.7, +40.6]}} & $-0.9$ {\scriptsize[-2.6, +0.8]}\\
 & 0.6 & \gain{20}{$+22.6$ {\scriptsize[+19.6, +25.5]}} & \gain{20}{$+21.6$ {\scriptsize[+18.8, +24.4]}} & $-1.0$ {\scriptsize[-2.7, +0.7]}\\
 & 1.0 & $-1.4$ {\scriptsize[-3.5, +0.6]} & $+1.5$ {\scriptsize[-0.2, +3.2]} & \gain{20}{$+2.9$ {\scriptsize[+1.3, +4.5]}}\\
\addlinespace
AIME & 0.3 & $+1.2$ {\scriptsize[-0.9, +3.3]} & $+0.6$ {\scriptsize[-1.4, +2.5]} & $-0.6$ {\scriptsize[-1.5, +0.1]}\\
 & 0.6 & $+0.7$ {\scriptsize[-1.5, +2.5]} & $+0.4$ {\scriptsize[-1.7, +2.0]} & $-0.3$ {\scriptsize[-0.9, +0.3]}\\
 & 1.0 & $+1.0$ {\scriptsize[+0.1, +2.3]} & $+1.5$ {\scriptsize[+0.3, +3.2]} & $+0.5$ {\scriptsize[-0.4, +1.5]}\\
\addlinespace
GPQA & 0.3 & \gain{20}{$+18.3$ {\scriptsize[+14.2, +22.5]}} & \gain{20}{$+19.4$ {\scriptsize[+15.3, +23.5]}} & $+1.1$ {\scriptsize[-1.0, +3.2]}\\
 & 0.6 & \gain{20}{$+14.2$ {\scriptsize[+10.3, +18.1]}} & \gain{20}{$+13.1$ {\scriptsize[+9.4, +16.9]}} & $-1.1$ {\scriptsize[-3.1, +0.9]}\\
 & 1.0 & $+0.6$ {\scriptsize[-2.5, +3.7]} & $+3.7$ {\scriptsize[+1.4, +6.0]} & $+3.1$ {\scriptsize[+0.1, +6.1]}\\
\bottomrule
\end{tabular*}
\caption{GPTQ3 accuracy differences in percentage points with 95\% intervals from a paired bootstrap over questions (Appendix~\ref{app:behavior-protocol}). Per-step is forward-KL per-step TempQuant. Green marks gains that remain significant after Holm correction over all 75 tests, which cover every task, temperature, and training loss; the table shows the comparisons with global and per-step TempQuant. Differences use paired attempts, so they can differ slightly from Table~\ref{tab:gptq3-main}.}
\label{tab:gptq3-global-significance}
\end{table}

\paragraph{Strict grading.}
A response that hits the length limit is cut off before it finishes. Our
graders take the last answer that appears in the text, so a response that
reached the right answer and then kept going, often repeating itself until
the limit, can still be graded correct. At $\tau=0.3$, for example, raw
GPTQ3 hits the limit on 94\% of GSM8K responses yet scores 18\%. To check
that TempQuant's gains do not come from grading unfinished responses, Table~\ref{tab:gptq3-strict} counts
every response that hits the length limit as wrong. The gains from TempQuant at $\tau=0.3$ and $0.6$ hold under this
stricter grading, and much of them comes from more responses finishing:
at $\tau=0.3$, GPTQ3 finishes few responses, while TempQuant finishes most
of its GSM8K and GPQA responses and about half of its MATH responses. Among
the responses that finish, GPTQ3 is more accurate, because it only finishes
its easiest ones. At $\tau=1.0$, global TempQuant finishes fewer GSM8K
responses than GPTQ3, so it falls further behind under strict grading,
while per-step TempQuant stays level or ahead.

\begin{table}[!htbp]
\centering\small
\setlength{\tabcolsep}{2.5pt}
\begin{tabular*}{\linewidth}{@{\extracolsep{\fill}}llrrrrrrrrr@{}}
\toprule
 & & \multicolumn{3}{c}{GSM8K} & \multicolumn{3}{c}{MATH} & \multicolumn{3}{c}{GPQA}\\
\cmidrule(lr){3-5}\cmidrule(lr){6-8}\cmidrule(lr){9-11}
$\tau$ & Policy & Strict & Fin.\ acc. & Fin.\ & Strict & Fin.\ acc. & Fin.\ & Strict & Fin.\ acc. & Fin.\\
\midrule
0.3 & BF16 & 94.28 & 94.7 & 99.6 & 94.90 & 96.5 & 98.7 & 38.83 & 41.9 & 92.7\\
0.3 & GPTQ3 & 5.46 & 92.3 & 5.9 & 2.67 & 96.4 & 2.8 & 0.06 & 7.7 & 0.8\\
0.3 & Global TempQuant & 66.22 & 88.1 & 75.2 & 39.12 & 75.7 & 51.8 & 22.35 & 26.2 & 85.2\\
0.3 & Per-step TempQuant (FKL) & 68.43 & 87.7 & 78.1 & 40.14 & 77.7 & 51.7 & 23.30 & 28.9 & 80.5\\
\midrule
0.6 & BF16 & 94.83 & 94.9 & 99.9 & 95.65 & 96.4 & 99.5 & 41.29 & 41.7 & 99.1\\
0.6 & GPTQ3 & 36.64 & 90.1 & 40.7 & 14.30 & 92.6 & 15.4 & 3.47 & 38.5 & 9.0\\
0.6 & Global TempQuant & 67.48 & 87.0 & 77.6 & 41.40 & 78.2 & 53.0 & 22.92 & 27.0 & 84.8\\
0.6 & Per-step TempQuant (FKL) & 65.19 & 86.9 & 75.0 & 36.85 & 78.9 & 46.8 & 20.83 & 27.8 & 75.0\\
\midrule
1.0 & BF16 & 94.92 & 95.0 & 99.9 & 95.05 & 95.5 & 99.8 & 43.50 & 43.6 & 99.8\\
1.0 & GPTQ3 & 69.86 & 87.4 & 79.9 & 40.95 & 81.9 & 50.1 & 17.74 & 26.3 & 67.5\\
1.0 & Global TempQuant & 61.83 & 85.7 & 72.1 & 37.45 & 70.5 & 53.1 & 17.23 & 23.9 & 72.2\\
1.0 & Per-step TempQuant (FKL) & 69.97 & 87.7 & 79.8 & 39.05 & 76.2 & 51.3 & 23.45 & 27.6 & 85.1\\
\bottomrule
\end{tabular*}
\caption{Qwen3-4B GPTQ3 under strict grading. Strict counts responses that hit the length limit as wrong; Fin.\ acc.\ is accuracy among responses that finished; Fin.\ is the share of responses that finished (\%). Per-step TempQuant is forward KL, two fits averaged.}
\label{tab:gptq3-strict}
\end{table}

\section{Additional experimental results}
\label{app:additional-results}

\subsection{Full results for Section~\ref{sec:final-evaluation}}
\label{app:full-results}
\begin{table}[!htbp]
\centering\small
\setlength{\tabcolsep}{4pt}
\begin{tabular*}{\linewidth}{@{\extracolsep{\fill}}llrrrrr@{}}
\toprule
Model & Policy & GSM8K & MATH & GPQA & \shortstack{At length limit (\%)\\GPTQ3 $\to$ global} & \shortstack{Forward KL\\removed (\%)}\\
\midrule
Qwen3-4B & BF16 & 94.83 & 95.75 & 41.48 & & \\
 & GPTQ3 & 50.36 & 29.90 & 10.98 & \smash{\raisebox{-1.4ex}{76.7 $\to$ 29.9}} & \smash{\raisebox{-1.4ex}{73}}\\
 & Global TempQuant & \gain{20}{80.80} & \gain{20}{52.35} & \gain{20}{25.19} &  & \\
 & Per-step TempQuant & \gain{20}{79.68} & \gain{20}{51.35} & \gain{20}{24.08} &  & \\
\midrule
Qwen3-8B & BF16 & 96.50 & 95.75 & 42.68 & & \\
 & GPTQ3 & 88.67 & 60.83 & 22.22 & \smash{\raisebox{-1.4ex}{29.2 $\to$ 11.7}} & \smash{\raisebox{-1.4ex}{70}}\\
 & Global TempQuant & \gain{20}{92.75} & \gain{20}{70.42} & \gain{20}{27.15} &  & \\
 & Per-step TempQuant & \gain{20}{92.92} & \gain{20}{69.33} & 25.51 &  & \\
\midrule
Gemma 4 31B & BF16 & 98.67 & 98.42 & 84.85 & & \\
 & GPTQ3 & 97.33 & 69.75 & 62.12 & \smash{\raisebox{-1.4ex}{17.1 $\to$ 1.3}} & \smash{\raisebox{-1.4ex}{79}}\\
 & Global TempQuant & 97.42 & \gain{20}{90.25} & \gain{20}{68.81} &  & \\
 & Per-step TempQuant & 97.00 & \gain{20}{89.67} & \gain{20}{70.08} &  & \\
\midrule
R1-Distill-Llama-8B & BF16 & 93.42 & 89.92 & 42.17 & & \\
 & GPTQ3 & 76.67 & 67.33 & 22.85 & \smash{\raisebox{-1.4ex}{5.0 $\to$ 0.3}} & \smash{\raisebox{-1.4ex}{66}}\\
 & Global TempQuant & 77.25 & 67.83 & 21.34 &  & \\
 & Per-step TempQuant & 76.83 & 67.50 & \gain{20}{26.58} &  & \\
\bottomrule
\end{tabular*}
\caption{3-bit models at $\tau=0.6$: accuracy (\%), the share of responses at the length limit before and after global TempQuant, and the forward KL it removes. Green marks gains over GPTQ3.}
\label{tab:cross-model}
\end{table}

In Table~\ref{tab:cross-model}, Qwen3-4B is evaluated on the full
benchmarks and the other models on 300 GSM8K, 300 MATH, and 198 GPQA
questions with four samples each. On Qwen3-8B and R1-Distill-Llama-8B, the
per-token oracle removes a similar share of the divergence as on Qwen3-4B
(Table~\ref{tab:new-families}).

\begin{table}[!htbp]
\centering\small
\setlength{\tabcolsep}{3.5pt}
\begin{tabular*}{\linewidth}{@{\extracolsep{\fill}}llrrrr@{}}
\toprule
Model & Policy & Forward KL & Reverse KL & JSD & TV\\
\midrule
Qwen3-8B & GPTQ3 & .915 & 1.310 & .0530 & .119\\
 & Global TempQuant $(0.95, 1)$ & .275 & 1.860 & .0590 & .141\\
 & Per-step TempQuant, forward KL & .424 & 1.922 & .0592 & .144\\
 & Per-step TempQuant, JSD & .874 & 1.353 & .0531 & .121\\
 & Per-step TempQuant, reverse KL & 1.538 & 1.122 & .0583 & .120\\
 & Oracle per-step TempQuant ($T$ only) & .162 & 1.073 & .0363 & .086\\
 & Oracle per-step TempQuant & \textbf{.143} & \textbf{.996} & \textbf{.0341} & \textbf{.081}\\
\midrule
R1-Distill-Llama-8B & GPTQ3 & .706 & .837 & .0477 & .125\\
 & Global TempQuant $(0.72, 1)$ & .242 & 1.227 & .0518 & .137\\
 & Per-step TempQuant, forward KL & .329 & 1.389 & .0532 & .143\\
 & Per-step TempQuant, JSD & .641 & .882 & .0476 & .126\\
 & Per-step TempQuant, reverse KL & 1.476 & .679 & .0574 & .136\\
 & Oracle per-step TempQuant ($T$ only) & .176 & .856 & .0370 & .100\\
 & Oracle per-step TempQuant & .153 & .757 & .0342 & .092\\
\midrule
\multicolumn{6}{@{}l}{\textit{Mean entropy before top-$p$ (nats) and the BF16 temperature with the same mean entropy}}\\[2pt]
\multicolumn{2}{@{}l}{Qwen3-8B} & \multicolumn{4}{r}{BF16 $.174$, GPTQ3 $.229$; BF16 at $T=0.78$}\\
\bottomrule
\end{tabular*}
\caption{Two further models quantized with GPTQ3, at requested temperature $\tau=0.6$, on held-out steps. Global TempQuant settings are searched on training steps and given as $(T,p)$; per-step values average two training runs. On Qwen3-4B, the gap between per-step and global TempQuant comes from the learned cutoff (Appendix~\ref{app:joint-oracle}).}
\label{tab:new-families}
\end{table}

\subsection{Details for Figure~\ref{fig:three-bit-context}}
\label{app:entropy-same-prefix}
\label{app:e14-entropy}
\label{app:three-bit-entropy}

\paragraph{The effective temperature.}
Figure~\ref{fig:three-bit-context}a reports, for each requested temperature
$\tau$, the BF16 temperature $T_{\rm eq}(\tau)$ whose mean entropy matches
the quantized model's on the same prefixes:
\begin{equation}
 \sum_i w_i H\!\left(\softmax\!\left(\frac{u_i}{T_{\rm eq}(\tau)}\right)\right)
 =\sum_i w_i H\!\left(\softmax\!\left(\frac{z_i}{\tau}\right)\right).
 \label{eq:entropy-equivalent-temperature}
\end{equation}
We use the 198 BF16-generated GPQA-Diamond responses, sampled at $0.6$ with
top-$p$ $.95$, with questions weighted equally. GPTQ3 is hotter than BF16 at
all three requested temperatures, on these responses and on MXFP4-generated
ones.

\paragraph{Random-noise control.}
\label{app:noise-temperature}
For the noise line, we add Gaussian noise to BF16's logits at every token,
with the same size as GPTQ3's difference from BF16 at that token, and find
the BF16 temperature with the same mean entropy. The noise gives $0.29$,
$0.58$, and $0.99$, against $0.36$, $0.74$, and $1.26$ for GPTQ3, with
question-level 95\% intervals narrower than $\pm0.02$.

\paragraph{Placing MXFP4.}
MXFP4 was measured on its own set of BF16-generated GPQA responses, where
BF16's entropy is known at only the three requested temperatures. We take
its mean entropy relative to BF16's on that text at each requested
temperature and find the BF16 temperature with the same ratio on the GPTQ3
responses. Applied to GPTQ3, the same procedure reproduces its values in the
figure exactly.

\begin{figure}[!htbp]
\centering
\includegraphics[width=\linewidth]{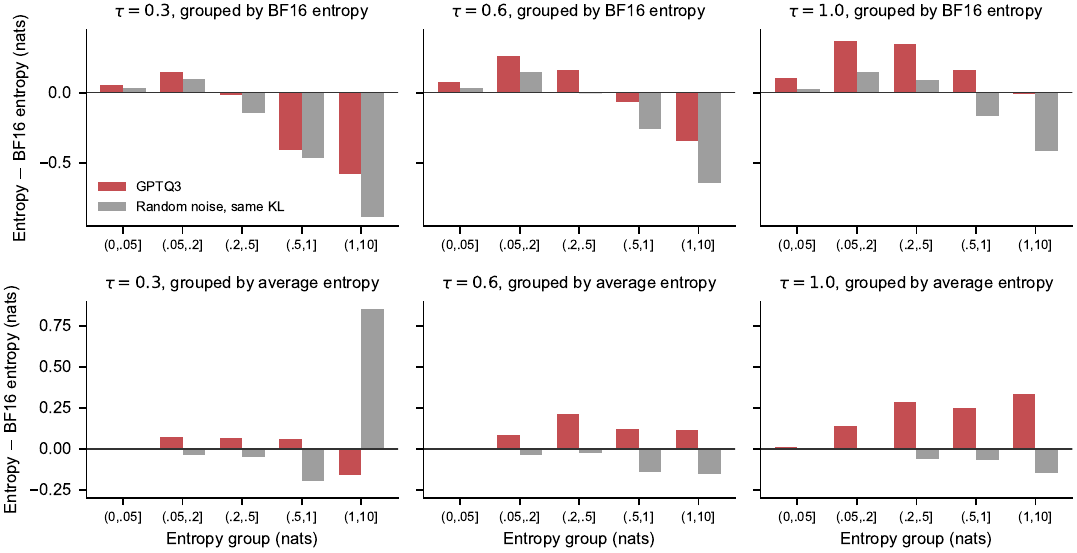}
\caption{Entropy change from BF16 for GPTQ3 and for random noise with the same forward KL, grouped by BF16 entropy (top) or by the average of the two entropies (bottom).}
\label{fig:entropy-regroup}
\end{figure}

\paragraph{Grouping in panel (b).}
Panel (b) groups tokens by the average entropy of the two models.
Figure~\ref{fig:entropy-regroup} shows why. Grouped by BF16's entropy alone,
both GPTQ3 and random noise with the same forward KL look flatter on
confident tokens and sharper on uncertain ones, since tokens picked for high
BF16 entropy tend to have lower entropy under any other model. Grouped by
the average of the two entropies, the noise is sharper and GPTQ3 flatter
almost everywhere.

\begin{figure}[!htbp]
\centering
\includegraphics[width=.85\linewidth]{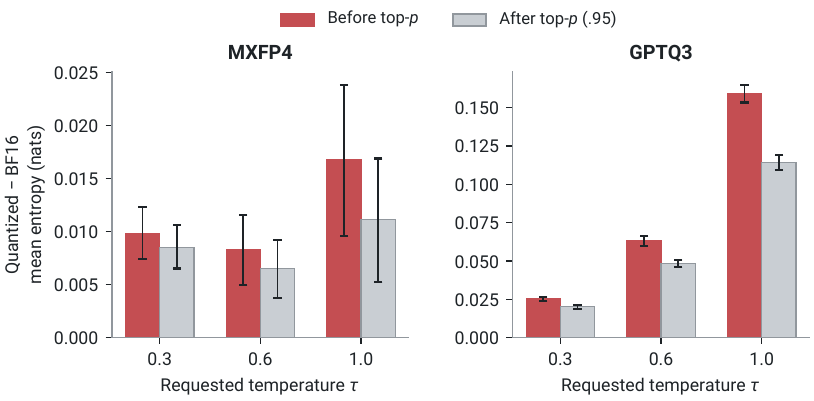}
\caption{Quantized-minus-BF16 mean entropy on BF16 responses before and after top-$p$ at $.95$, with 95\% intervals.}
\label{fig:multitemperature-entropy}
\label{fig:quantization-thermostat}
\end{figure}

\paragraph{Top-$p$ and entropy.}
Figure~\ref{fig:multitemperature-entropy} shows how the requested top-$p$
changes the gap. Top-$p$ removes the least likely tokens, which lowers the
entropy of both models and also narrows the gap between them: after
top-$p$, the gap from BF16 is 13--34\% smaller. Top-$p$ alone does not
remove it, so the quantized model still runs hotter than BF16 at the user's
settings.

\subsection{Global TempQuant}
\label{app:learned-global}

\paragraph{How the setting is fit.}
For each requested temperature, global TempQuant tries a coarse grid of
temperatures from $0.15$ to $3.5$ and top-$p$ values from $.9$ to $1$ on
the training steps, refines around the best pair, and is scored on held-out
steps from other questions. Nearby settings give almost the same divergence.
The downstream runs at $0.3$ and $1.0$ use $(1.09,1)$ and $(1.22,1)$. The
runs at $0.6$ use $(1.01,1)$ from an earlier, coarser search, whose forward
KL is within $2\%$ of the refined $(1.13,1)$ that the divergence tables use.

\paragraph{Do our measurement choices create the improvement?}
Plain KL is infinite when top-$p$ cuts a token BF16 keeps, so we mix a tiny
uniform probability into both distributions (weight $10^{-4}$). The
uncorrected model still pays a large penalty for each cut token, and the
penalty grows as the weight shrinks, so part of its divergence comes from
this choice. Table~\ref{tab:learned-global-eps} varies the weight a
thousandfold. The uncorrected model's divergence changes with it, because it
cuts a BF16 token on about $15\%$ of steps, while global TempQuant turns
top-$p$ off, cuts nothing, and stays the same. TempQuant stays far below the
uncorrected model at every weight, so the smoothing changes how large the
gap looks, not whether it exists. The other requested temperatures behave
the same way.

\begin{table}[!htbp]
\centering\small
\setlength{\tabcolsep}{4pt}
\begin{tabular*}{\linewidth}{@{\extracolsep{\fill}}lrrrrr@{}}
\toprule
 & \multicolumn{4}{c}{Forward KL at smoothing weight} & BF16 token\\
\cmidrule(lr){2-5}
Policy ($\tau=0.6$) & $10^{-6}$ & $10^{-5}$ & $10^{-4}$ & $10^{-3}$ & cut (\%)\\
\midrule
Uncorrected & 1.793 & 1.643 & 1.493 & 1.342 & 14.5\\
Global TempQuant (top-$p$ only) & 0.548 & 0.548 & 0.548 & 0.547 & 0.0\\
Global TempQuant & \textbf{0.397} & \textbf{0.397} & \textbf{0.397} & \textbf{0.396} & 0.0\\
Per-step TempQuant ($T$ only) & 0.546 & 0.528 & 0.510 & 0.491 & 2.8\\
Per-step TempQuant & 0.513 & 0.498 & 0.483 & 0.468 & 2.3\\
\bottomrule
\end{tabular*}
\caption{GPTQ3 forward KL to BF16 at $\tau=0.6$ for each smoothing weight, and the share of steps where the policy's top-$p$ cuts a token BF16 keeps.}
\label{tab:learned-global-eps}
\label{tab:learned-global-eps-03}
\label{tab:learned-global-eps-06}
\label{tab:learned-global-eps-10}
\end{table}

Our target is also BF16 as the user samples it, with top-$p$ cutting its
least likely tokens. To check that this choice does not create the gains,
Table~\ref{tab:untruncated-kl} compares against BF16's full distribution
instead.

\begin{table}[!htbp]
\centering\small
\setlength{\tabcolsep}{3pt}
\begin{tabular*}{\linewidth}{@{\extracolsep{\fill}}lrrrrrr@{}}
\toprule
 & \multicolumn{2}{c}{$\tau=0.3$} & \multicolumn{2}{c}{$\tau=0.6$} & \multicolumn{2}{c}{$\tau=1.0$}\\
\cmidrule(lr){2-3}\cmidrule(lr){4-5}\cmidrule(l){6-7}
Removed (\%) & Top-$p$ & Full & Top-$p$ & Full & Top-$p$ & Full\\
\midrule
Global TempQuant & 81 & 82 & 73 & 75 & 60 & 65\\
Per-step TempQuant & 75 & 76 & 68 & 69 & 45 & 49\\
Oracle per-step TempQuant & 89 & 89 & 84 & 84 & 75 & 76\\
\bottomrule
\end{tabular*}
\caption{Share of GPTQ3's uncorrected forward KL removed, measured against BF16 with top-$p$ $.95$ (Top-$p$), as in the rest of the paper, and against the untruncated BF16 distribution (Full).}
\label{tab:untruncated-kl}
\end{table}

\paragraph{What drives the fitted temperature up?}
Section~\ref{sec:which-way} argues that the flipped tokens push a single
temperature up. Table~\ref{tab:daca-ablation} tests this by fitting global
TempQuant only on tokens whose rankings are preserved.

\begin{table}[!htbp]
\centering\small
\setlength{\tabcolsep}{4pt}
\begin{tabular*}{\linewidth}{@{\extracolsep{\fill}}llrrrr@{}}
\toprule
$\tau$ & Fit on & Chosen $(T,p)$ & All tokens & Preserved & Flipped\\
\midrule
0.3 & All tokens & $(1.09,1)$ & \textbf{0.451} & 0.156 & \textbf{1.875}\\
 & Preserved only & $(0.45,1)$ & 0.733 & \textbf{0.122} & 3.676\\
0.6 & All tokens & $(1.13,1)$ & \textbf{0.397} & 0.151 & \textbf{1.579}\\
 & Preserved only & $(0.63,1)$ & 0.527 & \textbf{0.129} & 2.445\\
1.0 & All tokens & $(1.22,1)$ & \textbf{0.346} & 0.148 & \textbf{1.298}\\
 & Preserved only & $(0.85,1)$ & 0.402 & \textbf{0.136} & 1.684\\
\bottomrule
\end{tabular*}
\caption{Global TempQuant fit on all tokens or only on tokens whose rankings are preserved. Held-out GPTQ3 forward KL to BF16 on 768 steps, overall and split by rankings. Bold marks the lower value in each pair.}
\label{tab:daca-ablation}
\end{table}

\begin{figure}[!htbp]
\centering
\includegraphics[width=\linewidth]{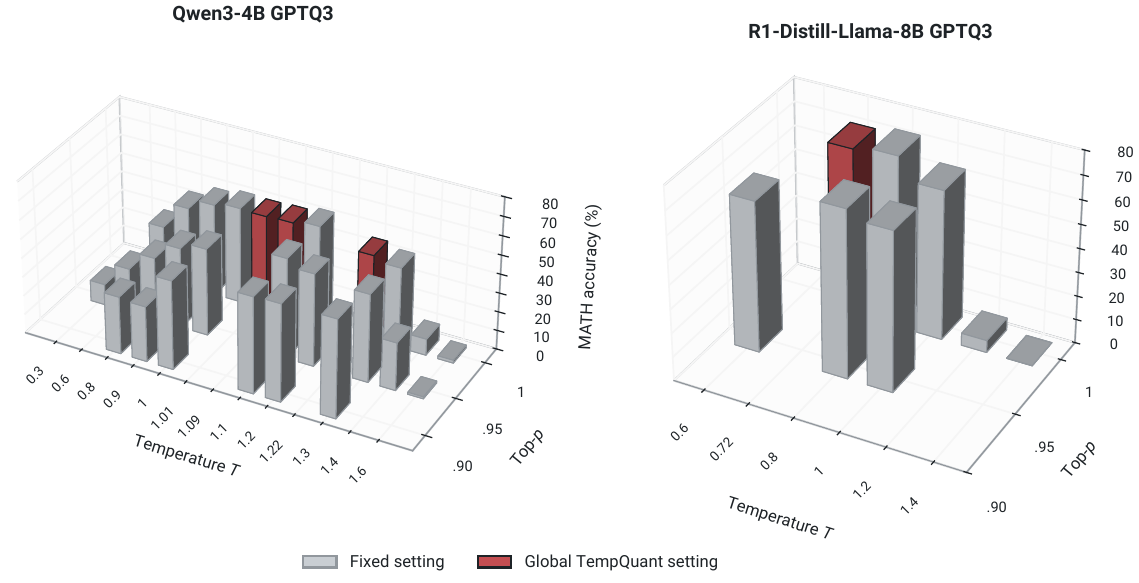}
\caption{MATH accuracy of GPTQ3 at fixed temperature and top-$p$ settings, and at the settings global TempQuant chooses from BF16 without answers, on the 198 gradable of 200 MATH questions with one sample each. For Qwen3-4B, $(1.0,1)$ shows the mean of two runs (52.0 and 49.0).}
\label{fig:label-free-3d}
\label{tab:label-free}
\end{figure}

\paragraph{Global TempQuant against tuning on answers.}
Global TempQuant needs no labels, while tuning the sampler on answers took
about 7,800 graded responses (Figure~\ref{fig:label-free-3d}). One could tune
on a small labeled set instead. Table~\ref{tab:label-budget} simulates this:
we choose the most accurate setting on 10 to 100 random MATH questions and
score it on the rest, against global TempQuant on the same questions.
Table~\ref{tab:tuned-baseline} compares the settings tuned on all 200
questions with global TempQuant on new MATH questions, GSM8K, and GPQA. In
both, global TempQuant reaches about the accuracy of answer tuning without
any labels.

\begin{table}[!htbp]
\centering\small
\setlength{\tabcolsep}{6pt}
\begin{tabular*}{\linewidth}{@{\extracolsep{\fill}}lrrr@{}}
\toprule
 & \multicolumn{3}{c}{Tuned minus global TempQuant (accuracy points)}\\
\cmidrule(l){2-4}
Labeled questions & Mean & 10th--90th percentile & Worse by 3+ points (\%)\\
\midrule
10 & +0.4 & [$-$3.7, +3.3] & 16\\
20 & +0.8 & [$-$3.4, +3.2] & 12\\
50 & +1.0 & [$-$1.4, +3.2] & 6\\
100 & +0.7 & [$-$2.3, +3.8] & 7\\
\bottomrule
\end{tabular*}
\caption{Tuning GPTQ3's sampler on a small labeled MATH set at $\tau=0.6$ and scoring it on the remaining questions, minus global TempQuant on the same questions (4,000 random splits of 198 questions).}
\label{tab:label-budget}
\end{table}

\begin{table}[!htbp]
\centering\small
\setlength{\tabcolsep}{4pt}
\begin{tabular*}{\linewidth}{@{\extracolsep{\fill}}lrrrr@{}}
\toprule
Qwen3-4B GPTQ3 setting & Tuning MATH & New MATH & GSM8K & GPQA\\
\midrule
Global TempQuant, $\tau=0.6$, $(1.01,1)$ & 49.00 & 53.92 & 80.80 & 25.63\\
Global TempQuant, $\tau=0.3$, $(1.09,1)$ & 48.75 & 54.67 & 81.71 & 24.24\\
Tuned on answers $(1.0,1)$ & 47.75 & 55.83 & 81.12 & 25.63\\
Tuned on answers $(1.2,.95)$ & 49.38 & 54.33 & 80.40 & 24.75\\
\midrule
R1-Distill-Llama-8B GPTQ3 setting & Tuning MATH & New MATH & GSM8K & GPQA\\
\midrule
Default $(0.6,.95)$ & 64.48 & 69.50 & 76.57 & 22.73\\
Global TempQuant $(0.72,1)$ & 70.03 & 69.17 & 76.72 & 21.72\\
Tuned on answers $(0.8,1)$ & 71.55 & 67.17 & 76.42 & 27.27\\
\bottomrule
\end{tabular*}
\caption{Accuracy (\%) of settings tuned on MATH answers and of global TempQuant, on the 200 tuning questions, 300 new MATH questions, GSM8K, and GPQA. Global TempQuant's GSM8K and GPQA scores come from the original runs with four samples per question, and the tuned settings use two to four new samples.}
\label{tab:tuned-baseline}
\end{table}

\paragraph{Other ways to choose a setting.}
\label{app:entropy-rules}
\label{app:fixed-one}
We also tried 12 rules in the spirit of ReSET \citep{lee2026reset} that use
a low temperature below an entropy threshold and a higher one above it. The
best reaches $49.5\%$ on the tuning questions, against $53.0\%$ for the best
fixed setting. Sampling every quantized model at $1.0$ with top-$p$ off would
be simpler, since raw GPTQ3 already does much better at $1.0$ than at $0.6$
(Table~\ref{tab:gptq3-main}), but the best setting differs between models.
R1-Distill-Llama-8B collapses a little above $1.0$, and matching BF16 picks a
lower temperature for it, close to its most accurate setting
(Figure~\ref{fig:label-free-3d}), while Qwen's 4-bit AWQ release is best
matched near $0.7$ (Appendix~\ref{app:awq}).

\subsection{Per-step examples}
\label{app:per-step-examples}
Table~\ref{tab:per-step-examples} shows two written prompts. In the first,
GPTQ3 puts all of its probability on ``multiple'' where BF16 writes
``common'', and per-step TempQuant raises the temperature far above the
global one, doubling the probability of ``common''. In the second, GPTQ3
already mostly agrees with BF16, and per-step TempQuant stays close to the
requested temperature, while global TempQuant's higher temperature moves
probability onto other tokens.

\begin{table}[!htbp]
\centering\small
\setlength{\tabcolsep}{7pt}
\begin{tabular}{@{}lcccc@{}}
\toprule
Text so far & \multicolumn{2}{p{4.6cm}}{\centering\emph{``\dots if a number is divisible by both 6 and 10, that means it has to be a''}}
& \multicolumn{2}{p{4.6cm}}{\centering\emph{``\dots if the number is 1, then it has 1 divisor, but 1 isn''}}\\
\addlinespace[2pt]
BF16's next token & \multicolumn{2}{c}{\fbox{\strut common}} & \multicolumn{2}{c}{\fbox{\strut 't}}\\
\cmidrule(lr){2-3}\cmidrule(l){4-5}
Policy & $T$ & $P(\text{common})$ & $T$ & $P(\text{'t})$\\
\midrule
BF16 & 0.60 & 1.00 & 0.60 & 1.00\\
GPTQ3 & 0.60 & 0.00 & 0.60 & 0.79\\
Global TempQuant & 1.13 & 0.14 & 1.13 & 0.18\\
Per-step TempQuant & 2.19 & 0.29 & 0.64 & 0.73\\
Oracle per-step TempQuant & 2.94 & 0.34 & 0.45 & 1.00\\
\bottomrule
\end{tabular}
\caption{GPTQ3: temperature each policy uses and the probability it gives BF16's next token at $\tau=0.6$, on written prompts.}
\label{tab:per-step-examples}
\end{table}

\subsection{Comparing training losses}
\label{app:objective-comparison}
\label{sec:objective-results}

Lowering the divergence a model is trained on need not lower the others.
Table~\ref{tab:four-divergence} measures every training loss on all four
divergences for GPTQ3. Each loss lowers its own divergence, and none lowers
all four at every temperature, which is why we report all three losses
instead of treating one divergence as the measure of fidelity.
The same holds for fixed settings. Among 140 fixed temperature and top-$p$
settings, none lowers all four on the training steps for GPTQ3 at $0.3$ and
$0.6$ or for MXFP4 at any temperature, and the few that do for GPTQ3 at $1.0$
lower them by at most 11\%.

Since per-step TempQuant is a learned predictor, a result could depend on a
lucky training run. Table~\ref{tab:seed-spread} retrains each loss five times
with different seeds: the divergences vary by only a few percent between
runs, for example from $0.460$ to $0.483$ for forward KL at $0.6$, so a
single run is representative. Unless noted, downstream results average two
runs.

\begin{table}[!htbp]
\centering\small
\setlength{\tabcolsep}{4pt}
\begin{tabular*}{\linewidth}{@{\extracolsep{\fill}}lrrrr@{}}
\toprule
Policy & Forward KL & Reverse KL & JSD & Total variation\\
\midrule
\multicolumn{5}{@{}l}{\textit{Temperature ($\tau$) = 0.3}}\\[2pt]
Global TempQuant (forward KL) & \textbf{-81} & +36 & +4 & +27\\
Global TempQuant (JSD) & \textbf{-48} & +7 & \textbf{-6} & +5\\
Global TempQuant (reverse KL) & +26 & \textbf{-1} & +5 & 0\\
Global TempQuant (top-$p$ only, forward KL) & \textbf{-57} & 0 & \textbf{-2} & 0\\
Per-step TempQuant (forward KL) & \textbf{-75} & +33 & +2 & +25\\
Per-step TempQuant (JSD) & \textbf{-43} & +7 & \textbf{-6} & +4\\
Per-step TempQuant (reverse KL) & +23 & \textbf{-1} & +4 & 0\\
\midrule
\multicolumn{5}{@{}l}{\textit{Temperature ($\tau$) = 0.6}}\\[2pt]
Global TempQuant (forward KL) & \textbf{-73} & +40 & +9 & +20\\
Global TempQuant (JSD) & \textbf{-40} & +6 & \textbf{-2} & +2\\
Global TempQuant (reverse KL) & +114 & \textbf{-9} & +20 & +4\\
Global TempQuant (top-$p$ only, forward KL) & \textbf{-63} & +3 & \textbf{-1} & +1\\
Per-step TempQuant (forward KL) & \textbf{-68} & +56 & +14 & +29\\
Per-step TempQuant (JSD) & \textbf{-17} & +5 & \textbf{-2} & +1\\
Per-step TempQuant (reverse KL) & +62 & \textbf{-7} & +9 & 0\\
\midrule
\multicolumn{5}{@{}l}{\textit{Temperature ($\tau$) = 1.0}}\\[2pt]
Global TempQuant (forward KL) & \textbf{-60} & +28 & +11 & +13\\
Global TempQuant (JSD) & \textbf{-2} & \textbf{-8} & 0 & \textbf{-1}\\
Global TempQuant (reverse KL) & +297 & \textbf{-24} & +40 & +12\\
Global TempQuant (top-$p$ only, forward KL) & \textbf{-58} & +8 & +3 & +4\\
Per-step TempQuant (forward KL) & \textbf{-45} & +35 & +12 & +15\\
Per-step TempQuant (JSD) & +8 & \textbf{-4} & 0 & \textbf{-1}\\
Per-step TempQuant (reverse KL) & +124 & \textbf{-19} & +11 & +1\\
\bottomrule
\end{tabular*}
\caption{Change in GPTQ3 divergence from BF16 relative to uncorrected GPTQ3 (\%), on held-out steps. Bold marks reductions.}
\label{tab:four-divergence}
\end{table}

\begin{table}[!htbp]
\centering\small
\setlength{\tabcolsep}{3pt}
\begin{tabular*}{\linewidth}{@{\extracolsep{\fill}}llrrr@{}}
\toprule
Training loss & $\tau$ & Forward KL & Reverse KL & JSD\\
\midrule
Forward KL & 0.3 & \estci{0.594}{0.594}{0.594} & \estci{3.506}{3.500}{3.514} & \estci{0.1006}{0.1005}{0.1008}\\
 & 0.6 & \estci{0.470}{0.460}{0.483} & \estci{3.222}{3.212}{3.235} & \estci{0.0943}{0.0940}{0.0945}\\
 & 1.0 & \estci{0.484}{0.477}{0.491} & \estci{2.451}{2.425}{2.492} & \estci{0.0787}{0.0783}{0.0795}\\
\midrule
Reverse KL & 0.3 & \estci{2.939}{2.895}{2.987} & \estci{2.600}{2.598}{2.602} & \estci{0.1028}{0.1026}{0.1032}\\
 & 0.6 & \estci{2.395}{2.395}{2.395} & \estci{1.927}{1.927}{1.927} & \estci{0.0900}{0.0900}{0.0900}\\
 & 1.0 & \estci{1.940}{1.937}{1.941} & \estci{1.490}{1.489}{1.490} & \estci{0.0789}{0.0789}{0.0789}\\
\midrule
JSD & 0.3 & \estci{1.365}{1.364}{1.365} & \estci{2.801}{2.799}{2.803} & \estci{0.0925}{0.0924}{0.0925}\\
 & 0.6 & \estci{1.242}{1.221}{1.253} & \estci{2.171}{2.169}{2.173} & \estci{0.0810}{0.0809}{0.0811}\\
 & 1.0 & \estci{0.944}{0.940}{0.952} & \estci{1.780}{1.778}{1.781} & \estci{0.0711}{0.0711}{0.0712}\\
\bottomrule
\end{tabular*}
\caption{GPTQ3 divergence from BF16 for five per-step TempQuant training runs per loss: mean [minimum, maximum].}
\label{tab:seed-spread}
\end{table}

Table~\ref{tab:multitemperature-support} shows why the losses trade off.
Forward KL charges most for cutting a token BF16 would use, so the
forward-KL predictor warms and opens its cutoff, cutting almost none of
BF16's probability; the warming also moves probability onto tokens BF16's
own top-$p$ removes, which reverse KL and total variation charge for.
Reverse KL does the opposite: it cools and tightens the cutoff, placing the
least probability on tokens BF16 removes while cutting more of BF16's.
JSD lands in between.

\begin{table}[!htbp]
\centering\small
\setlength{\tabcolsep}{5pt}
\begin{tabular*}{\linewidth}{@{\extracolsep{\fill}}lrrrr@{}}
\toprule
 & \shortstack{BF16 probability\\the policy cuts (\%)} & \shortstack{Policy probability on\\tokens BF16 cuts (\%)} & \shortstack{Mean\\temperature} & \shortstack{Mean\\top-$p$}\\
\midrule
\multicolumn{5}{@{}l}{\textit{$\tau=0.3$}}\\[1pt]
GPTQ3 & 11.1 & 12.3 & 0.30 & 0.950\\
Forward KL & \textbf{1.0} & 18.1 & 1.09 & 0.975\\
JSD & 5.6 & 13.7 & 0.62 & 0.962\\
Reverse KL & 14.0 & \textbf{12.1} & 0.48 & 0.942\\
\midrule
\multicolumn{5}{@{}l}{\textit{$\tau=0.6$}}\\[1pt]
GPTQ3 & 6.5 & 9.8 & 0.60 & 0.950\\
Forward KL & \textbf{0.7} & 17.0 & 1.40 & 0.974\\
JSD & 5.1 & 10.5 & 0.68 & 0.958\\
Reverse KL & 11.5 & \textbf{8.7} & 0.46 & 0.929\\
\midrule
\multicolumn{5}{@{}l}{\textit{$\tau=1.0$}}\\[1pt]
GPTQ3 & 3.3 & 9.0 & 1.00 & 0.950\\
Forward KL & \textbf{0.9} & 13.1 & 1.37 & 0.974\\
JSD & 3.7 & 8.6 & 0.94 & 0.956\\
Reverse KL & 9.3 & \textbf{6.6} & 0.62 & 0.928\\
\bottomrule
\end{tabular*}
\caption{GPTQ3 and per-step TempQuant trained with each loss, on held-out steps: probability each side places on tokens the other's top-$p$ removes, and the mean settings the predictor chooses. Bold marks the lowest value in each mass column. Rows average two training runs.}
\label{tab:multitemperature-support}
\end{table}

\subsection{What learning top-\texorpdfstring{$p$}{p} adds}
\label{app:matched-controls}
\label{app:t060-extension}
Per-step TempQuant learns a temperature and a top-$p$ cutoff. To see what
the cutoff adds, we compare it with per-step TempQuant ($T$ only), trained
the same way but keeping $p=.95$. On divergence, learning the cutoff lowers
forward KL at every requested temperature, and all four divergences at $0.6$
and $1.0$ (Table~\ref{tab:topp-divergence}). On accuracy, the two versions
are close (Table~\ref{tab:topp-downstream}); the difference on GSM8K at
$0.6$ is $+0.7$ points with a 95\% interval of $[-0.1, +1.6]$. Global
TempQuant that learns only top-$p$ reaches much lower accuracy than global
TempQuant that also learns the temperature, a mean of $28.3\%$ against
$40.0\%$ at $0.6$. With its learned top-$p$ turned off at $1.0$, per-step
TempQuant scores $0.6$ points lower on GSM8K and $1.3$ on MATH, with
intervals $[-1.5, +0.2]$ and $[-3.4, +0.8]$. Most of the correction
therefore comes from the temperature.

\begin{table}[!htbp]
\centering\small
\setlength{\tabcolsep}{5pt}
\begin{tabular*}{\linewidth}{@{\extracolsep{\fill}}lrrrr@{}}
\toprule
Per-step TempQuant & Forward KL & Reverse KL & JSD & Total variation\\
\midrule
\multicolumn{5}{@{}l}{\textit{$\tau=0.3$}}\\[1pt]
Temperature only & $-$69.4 & +33.4 & \textbf{+1.7} & \textbf{+24.3}\\
Temperature and top-$p$ & \textbf{$-$75.4} & \textbf{+33.3} & +1.8 & +24.6\\
\midrule
\multicolumn{5}{@{}l}{\textit{$\tau=0.6$}}\\[1pt]
Temperature only & $-$65.9 & +64.3 & +18.5 & +33.2\\
Temperature and top-$p$ & \textbf{$-$67.7} & \textbf{+55.6} & \textbf{+14.5} & \textbf{+28.5}\\
\midrule
\multicolumn{5}{@{}l}{\textit{$\tau=1.0$}}\\[1pt]
Temperature only & $-$43.1 & +43.4 & +15.5 & +18.7\\
Temperature and top-$p$ & \textbf{$-$44.9} & \textbf{+34.8} & \textbf{+11.7} & \textbf{+14.8}\\
\bottomrule
\end{tabular*}
\caption{Change in GPTQ3 divergence from BF16 relative to uncorrected GPTQ3 (\%), on held-out steps, for forward-KL per-step TempQuant with and without a learned top-$p$. Bold marks the lower of each pair. Rows average two training runs.}
\label{tab:topp-divergence}
\label{tab:gptq3-core-dev}
\end{table}

\begin{table}[!htbp]
\centering\small
\setlength{\tabcolsep}{5pt}
\begin{tabular*}{\linewidth}{@{\extracolsep{\fill}}lrrrr@{}}
\toprule
Policy & GSM8K & MATH & AIME & GPQA\\
\midrule
\multicolumn{5}{@{}l}{\textit{Global TempQuant, $\tau=0.6$}}\\[1pt]
Top-$p$ only & 62.06 & 35.40 & 0.47 & 15.21\\
Temperature and top-$p$ & \textbf{80.80} & \textbf{52.35} & \textbf{1.72} & \textbf{25.19}\\
\midrule
\multicolumn{5}{@{}l}{\textit{Per-step TempQuant, $\tau=0.6$}}\\[1pt]
Temperature only & 78.93 & \textbf{51.91} & 1.25 & --\\
Temperature and top-$p$ & \textbf{79.68} & 51.35 & \textbf{1.43} & --\\
\midrule
\multicolumn{5}{@{}l}{\textit{Per-step TempQuant, $\tau=1.0$}}\\[1pt]
Learned top-$p$ turned off & 80.22 & 50.60 & -- & --\\
Learned top-$p$ on & \textbf{80.86} & \textbf{51.90} & -- & --\\
\bottomrule
\end{tabular*}
\caption{GPTQ3 accuracy (\%) with and without learning top-$p$. Per-step rows use forward KL and average two training runs; -- marks a comparison not run.}
\label{tab:topp-downstream}
\label{tab:t060-added-grades}
\label{tab:t060-added-contrasts}
\label{tab:p1-ablation}
\label{tab:global-vs-perstep}
\end{table}

\subsection{Behavior beyond accuracy}
\label{app:gptq3-full-downstream}
\label{app:behavior}

\paragraph{Closeness.}
The closeness score summarizes how far a policy's generated text is from
BF16's on four statistics: word overlap with BF16's responses to the same
question (BLEU against a different half of BF16's samples, so BF16 against
itself scores below 100); self-BLEU, which measures how alike the policy's
own responses to the same question are; the share of repeated four-grams;
and the share of responses that hit the length limit. For each task it is
$100-\frac14\sum_m|x_m-b_m|$, where $x_m$ are the policy's statistics and
$b_m$ BF16's, so BF16 scores 100; for reasoning we average it over GSM8K,
MATH, AIME, and GPQA. Table~\ref{tab:closeness-breakdown} shows the four
statistics at $\tau=0.6$. On reasoning, the gaps in repetition and in
responses at the length limit dominate, so the score mostly tracks looping.
On XSum, where nothing loops, word overlap and self-BLEU dominate, so it
tracks variety.
The closest controller in each block follows the split of
Section~\ref{sec:behavior-results}. Long reasoning under heavy degradation,
GPTQ3 on the four reasoning tasks, should favor forward KL, and the forward
KL controllers are closest. Short text should favor reverse KL, and the
reverse KL controller is closest on both GPTQ3 and MXFP4 summaries.
Reasoning under light degradation, MXFP4 on GPQA, should favor JSD, and the
JSD controller is the closest of the three.

\begin{table}[!htbp]
\centering\small
\setlength{\tabcolsep}{4pt}
\begin{tabular*}{\linewidth}{@{\extracolsep{\fill}}lrrrrr@{}}
\toprule
Policy ($\tau=0.6$) & \shortstack{Word overlap\\with BF16} & Self-BLEU & \shortstack{Repeated\\4-grams (\%)} & \shortstack{At length\\limit (\%)} & Closeness\\
\midrule
\multicolumn{6}{@{}l}{\textit{GPTQ3, reasoning (mean of four tasks)}}\\[1pt]
BF16 & 66.1 & 72.0 & 35.3 & 2.6 & 100\\
GPTQ3 & 7.3 & 55.6 & 89.6 & 83.7 & 47.3\\
Global TempQuant & \textbf{18.0} & \textbf{64.8} & 59.2 & \textbf{42.7} & \textbf{70.2}\\
Per-step TempQuant, forward KL & 16.6 & 62.6 & \textbf{58.6} & 48.2 & 68.1\\
Per-step TempQuant, JSD & 9.0 & 56.9 & 85.5 & 76.8 & 50.9\\
Per-step TempQuant, reverse KL & 4.7 & 55.1 & 95.8 & 96.9 & 41.7\\
\midrule
\multicolumn{6}{@{}l}{\textit{GPTQ3, XSum}}\\[1pt]
BF16 & 85.7 & 90.4 & 0.02 & 0.0 & 100\\
GPTQ3 & 41.5 & 80.7 & 0.30 & 0.2 & 86.4\\
Per-step TempQuant, forward KL & 37.1 & 54.3 & \textbf{0.06} & \textbf{0.0} & 78.8\\
Per-step TempQuant, JSD & 40.0 & 73.5 & 0.51 & 0.4 & 84.1\\
Per-step TempQuant, reverse KL & \textbf{42.0} & \textbf{87.7} & 0.25 & 0.1 & \textbf{88.3}\\
\midrule
\multicolumn{6}{@{}l}{\textit{MXFP4, GPQA}}\\[1pt]
BF16 & 69.2 & 75.8 & 42.8 & 0.7 & 100\\
MXFP4 & 56.8 & \textbf{74.4} & \textbf{44.1} & \textbf{0.4} & \textbf{96.2}\\
Per-step TempQuant, forward KL & 55.7 & 71.3 & 32.3 & 0.1 & 92.7\\
Per-step TempQuant, JSD & \textbf{57.2} & 73.9 & 40.7 & 0.3 & 95.9\\
Per-step TempQuant, reverse KL & 55.8 & \textbf{74.4} & 48.8 & 1.4 & 94.7\\
\midrule
\multicolumn{6}{@{}l}{\textit{MXFP4, XSum}}\\[1pt]
BF16 & 86.4 & 90.2 & 0.00 & 0.0 & 100\\
MXFP4 & 48.7 & 81.7 & 0.00 & 0.0 & 88.5\\
Per-step TempQuant, forward KL & 46.2 & 68.4 & 0.00 & 0.0 & 84.5\\
Per-step TempQuant, JSD & 48.1 & 77.6 & 0.01 & 0.0 & 87.3\\
Per-step TempQuant, reverse KL & \textbf{48.9} & \textbf{84.7} & 0.01 & 0.0 & \textbf{89.3}\\
\bottomrule
\end{tabular*}
\caption{The four statistics behind the closeness score at $\tau=0.6$. Closeness is $100$ minus the average absolute gap from BF16 over the four statistics, computed per task; for reasoning, the statistics and closeness are averaged over GSM8K, MATH, AIME, and GPQA. Bold marks the quantized value nearest BF16's.}
\label{tab:closeness-breakdown}
\end{table}

\paragraph{MAUVE.}
MAUVE \citep{pillutla2023mauve} compares the distribution of a policy's
responses with BF16's (Table~\ref{tab:mauve}). It is computed on 512
responses per task and temperature, with the same questions and seeds for
every policy and each response truncated to its first 1,024 GPT-2 tokens.
At $0.3$ and $0.6$, both forms of TempQuant score above raw GPTQ3 on every
task, and all policies stay below the agreement between two halves of
BF16's own samples.

\begin{table}[!htbp]
\centering\small
\setlength{\tabcolsep}{5pt}
\begin{tabular*}{\linewidth}{@{\extracolsep{\fill}}llrrrr@{}}
\toprule
Task & $\tau$ & BF16 vs BF16 & GPTQ3 & Global TempQuant & Per-step TempQuant\\
\midrule
MATH & 0.3 & .939 & .579 & .879 & \textbf{.883}\\
 & 0.6 & .967 & .797 & \textbf{.881} & .830\\
 & 1.0 & .983 & .898 & .851 & \textbf{.902}\\
\addlinespace
GSM8K & 0.3 & .951 & .458 & \textbf{.662} & .652\\
 & 0.6 & .958 & .485 & .582 & \textbf{.657}\\
 & 1.0 & .961 & .579 & \textbf{.637} & .538\\
\addlinespace
AIME & 0.3 & .958 & .548 & \textbf{.841} & .830\\
 & 0.6 & .946 & .806 & \textbf{.864} & .835\\
 & 1.0 & .973 & .776 & .752 & \textbf{.820}\\
\addlinespace
GPQA & 0.3 & .961 & .439 & .830 & \textbf{.881}\\
 & 0.6 & .974 & .731 & \textbf{.918} & .861\\
 & 1.0 & .942 & \textbf{.877} & .513 & .769\\
\bottomrule
\end{tabular*}
\caption{GPTQ3 MAUVE against BF16 (higher is closer). BF16 vs BF16 compares two halves of BF16's own samples; per-step TempQuant uses forward KL. Bold marks the quantized policy closest to BF16.}
\label{tab:mauve}
\end{table}

\subsection{Overthinking}
\label{app:overthinking}
Quantized reasoning models tend to overthink, sampling words such as
``Wait'' and ``Alternatively'' that start unnecessary rechecking
\citep{lotfi2026overthinking}. GPTQ3 does the same. Continuing generation
from 14 points where GPTQ3 strongly prefers ``Wait'' and BF16 does not,
chosen to show the effect where it is largest, raw GPTQ3 writes about twice
as many overthinking markers as BF16, and the learned settings barely change
this (Table~\ref{tab:overthinking}). Table~\ref{tab:overthinking-example}
shows one such point. GPTQ3 always opens with ``Wait'' there and BF16 never
does; per-step TempQuant warms sharply at this step and opens with doubt
less often than global TempQuant. Across all 14 points, the learned
settings and even the oracle move only part of the way to BF16. Changing the
temperature delays this behavior more than it removes it.

\begin{table}[!htbp]
\centering\small
\setlength{\tabcolsep}{4pt}
\begin{tabular*}{\linewidth}{@{\extracolsep{\fill}}lrp{0.47\linewidth}r@{}}
\toprule
\multicolumn{4}{@{}p{\linewidth}@{}}{\emph{Text so far: ``\dots So maybe I should start by finding the LCM of 6 and 10. Then, once I have that, I can figure out the smallest multiple of that LCM that's greater than 100.''}}\\
\midrule
Policy & $T$ & Start of one continuation & Opens with doubt\\
\midrule
BF16 & 0.60 & Alright, let me recall how to calculate the LCM. I think the formula is LCM$(a,b) = ab/\mathrm{GCD}(a,b)$\dots & 0/16\\
GPTQ3 & 0.60 & \textbf{Wait}, but let me check. The LCM of 6 and 10\dots{} \textbf{Hmm}, 6 and 10. Let me recall how to calculate the LCM\dots & 16/16\\
Global TempQuant & 1.13 & \textbf{Wait}, but let me check. Let me remember that LCM of 6 and 10. But 6 and 10 have a common factor\dots & 14/16\\
Per-step TempQuant & 1.91 & Alright, so let's start with finding the LCM of 6 and 10. To find the LCM of two numbers, I think I need to find their prime factors\dots & 10/16\\
Oracle per-step TempQuant & 2.58 & \textbf{Hmm}, right. Let me recall how to find the LCM. To find the LCM of two numbers, I can use the formula\dots & 9/16\\
\bottomrule
\end{tabular*}
\caption{GPTQ3 continuations of one written problem at $\tau=0.6$ from the same seed. Of 16 sampled continuations per policy, ``Opens with doubt'' counts those whose first word is ``Wait'', ``Hmm'', or ``Alternatively'' (bold), which start the rechecking \citet{lotfi2026overthinking} associate with overthinking.}
\label{tab:overthinking-example}
\end{table}

\begin{table}[!htbp]
\centering\small
\setlength{\tabcolsep}{6pt}
\begin{tabular*}{\linewidth}{@{\extracolsep{\fill}}lrrrr@{}}
\toprule
Policy & \shortstack{Starts with\\BF16's word (\%)} & \shortstack{Opens with\\doubt (\%)} & \shortstack{``Wait'' per\\continuation} & \shortstack{Markers per\\continuation}\\
\midrule
BF16 & 83 & 18 & 1.4 & 8.0\\
\midrule
Raw GPTQ3 & 5 & 75 & 5.4 & 18.4\\
Global TempQuant & 12 & 63 & 5.1 & 18.5\\
Per-step TempQuant & \textbf{13} & \textbf{60} & \textbf{5.0} & \textbf{17.5}\\
Oracle per-step TempQuant & 15 & 52 & 4.5 & 15.8\\
\bottomrule
\end{tabular*}
\caption{Continuations from 14 points where GPTQ3 prefers ``Wait'' and BF16 does not, 16 per point and policy, 400 tokens each, with matched seeds. Opens with doubt counts continuations whose first words are ``Wait'', ``Hmm'', or ``Alternatively''. Markers are the 50 overthinking markers of \citet{lotfi2026overthinking}. Bold marks the best non-oracle policy.}
\label{tab:overthinking}
\end{table}
\label{app:multitemperature-readout}

\subsection{MXFP4 results}
\label{app:mxfp4-results}
\label{app:math-results}
This section collects the MXFP4 versions of the GPTQ3 results.

\paragraph{Divergence.}
Global TempQuant removes about three quarters of MXFP4's forward KL at every
requested temperature (Table~\ref{tab:mxfp4-learned-global}). On MXFP4 at
$0.6$, the per-token oracle removes about four fifths of the forward KL, and
temperature alone about two thirds.

\begin{table}[!htbp]
\centering\small
\setlength{\tabcolsep}{6pt}
\begin{tabular*}{\linewidth}{@{\extracolsep{\fill}}lrrr@{}}
\toprule
Policy & $\tau=0.3$ & $\tau=0.6$ & $\tau=1.0$\\
\midrule
Uncorrected & 1.226 & 0.722 & 0.582\\
Top-$p$ off & 0.488 & 0.264 & 0.172\\
Global TempQuant ($T$ only) & 0.330 & 0.282 & 0.255\\
Global TempQuant & \textbf{0.236} & \textbf{0.189} & \textbf{0.153}\\
\midrule
Per-step TempQuant & 0.330 & 0.246 & 0.211\\
\bottomrule
\end{tabular*}
\caption{MXFP4 version of Table~\ref{tab:fitted-control-summary}: forward KL to BF16 on held-out steps. Global TempQuant uses $(T,p)=(1.21,1)$, $(1.27,1)$, and $(1.43,1)$. These MXFP4 next-token distributions come from a separate collection on the same steps. On GPTQ3, the gap between per-step and global TempQuant comes from the learned cutoff (Appendix~\ref{app:joint-oracle}).}
\label{tab:mxfp4-learned-global}
\end{table}

\paragraph{Accuracy.}
Table~\ref{tab:objective-benchmarks} is the MXFP4 version of
Table~\ref{tab:gptq3-main}. MXFP4's closeness on GPQA is in
Table~\ref{tab:closeness-breakdown}.

\begin{table}[!htbp]
\centering\small
\setlength{\tabcolsep}{3.5pt}
\renewcommand{\arraystretch}{1.08}
\begin{tabular*}{\linewidth}{@{\extracolsep{\fill}}lrrrrrr@{}}
\toprule
 & \multicolumn{4}{c}{Accuracy (\%)} & Mean & At length\\
\cmidrule(lr){2-5}
Policy & GSM8K & MATH & AIME & GPQA & accuracy & limit (\%)\\
\midrule
\multicolumn{7}{@{}l}{\textit{Temperature ($\tau$) = 0.3}}\\[1pt]
BF16 & 94.47 & 95.40 & 65.6 & 39.8 & 73.82 & 5.6\\
\cmidrule(lr){1-7}
MXFP4 & 93.67 & 91.95 & 43.2 & \textbf{41.5} & 67.58 & 3.5\\
Global TempQuant & \textbf{93.76} & \textbf{93.20} & 47.6 & 40.8 & \textbf{68.85} & \textbf{1.8}\\
Per-step TempQuant, forward KL & 93.75 & 92.48 & \gain{20}{\textbf{47.9}} & 40.1 & 68.56 & 1.8\\
Per-step TempQuant, JSD & 93.56 & 92.70 & 46.0 & 39.5 & 67.95 & 2.4\\
Per-step TempQuant, reverse KL & 93.51 & 92.15 & 44.1 & 39.4 & 67.30 & 3.9\\
\midrule
\multicolumn{7}{@{}l}{\textit{Temperature ($\tau$) = 0.6}}\\[1pt]
BF16 & 94.47 & 95.55 & 68.2 & 41.5 & 74.95 & 2.6\\
\cmidrule(lr){1-7}
MXFP4 & 93.27 & 92.45 & 45.8 & 39.3 & 67.71 & 3.1\\
Global TempQuant & \textbf{93.93} & 92.50 & \textbf{48.1} & 39.3 & 68.47 & \textbf{1.7}\\
Per-step TempQuant, forward KL & 93.67 & \textbf{93.10} & 46.9 & \textbf{42.0} & \textbf{68.94} & 2.0\\
Per-step TempQuant, JSD & 93.66 & 93.08 & 48.0 & 40.7 & 68.87 & 2.6\\
Per-step TempQuant, reverse KL & 93.18 & 92.83 & 44.5 & 39.5 & 67.50 & 4.2\\
\midrule
\multicolumn{7}{@{}l}{\textit{Temperature ($\tau$) = 1.0}}\\[1pt]
BF16 & 95.02 & 95.55 & 69.3 & 42.7 & 75.64 & 2.3\\
\cmidrule(lr){1-7}
MXFP4 & 93.65 & 92.55 & 48.1 & 39.6 & 68.48 & 2.2\\
Global TempQuant & 93.48 & 92.45 & 43.5 & 36.8 & 66.57 & 5.9\\
Per-step TempQuant, forward KL & \textbf{93.91} & \textbf{93.20} & 46.6 & \textbf{40.9} & 68.65 & \textbf{2.0}\\
Per-step TempQuant, JSD & 93.88 & 92.68 & 48.2 & \textbf{40.9} & \textbf{68.92} & 2.2\\
Per-step TempQuant, reverse KL & 93.48 & 93.15 & \textbf{48.8} & 39.3 & 68.67 & 3.0\\
\bottomrule
\end{tabular*}
\caption{MXFP4 version of Table~\ref{tab:gptq3-main}: accuracy, and the share of responses that hit the length limit, averaged over the four tasks. Bold marks the best quantized value (lowest for the length limit); green marks gains over MXFP4 that remain significant after Holm correction (Appendix~\ref{app:multiplicity}).}
\label{tab:objective-benchmarks}
\end{table}

\paragraph{Summaries.}
\label{app:xsum-behavior}
Table~\ref{tab:mxfp4-xsum} is the MXFP4 version of Table~\ref{tab:xsum-diversity},
with variety measured by self-BLEU, how alike a policy's summaries of the
same article are. The reduction in the reverse KL controller's
self-BLEU gap from BF16 has a 95\% interval below zero at each temperature.

\begin{table}[!htbp]
\centering\small
\setlength{\tabcolsep}{6pt}
\renewcommand{\arraystretch}{1.05}
\begin{tabular*}{\linewidth}{@{\extracolsep{\fill}}lrrr@{}}
\toprule
Policy & Self-BLEU & ROUGE-2 & Closeness\\
\midrule
\multicolumn{4}{@{}l}{\textit{Temperature ($\tau$) = 0.3}}\\[1pt]
BF16 (value) & 95.4 & 6.76 & 100\\
\cmidrule(lr){1-4}
$\Delta$ MXFP4 & $-$8.6 & \textbf{$-$1.14} & $-$14.0\\
$\Delta$ Per-step TempQuant, forward KL & $-$22.4 & $-$1.20 & $-$17.7\\
$\Delta$ Per-step TempQuant, JSD & $-$13.6 & $-$1.16 & $-$15.2\\
$\Delta$ Per-step TempQuant, reverse KL & \textbf{$-$6.9} & $-$1.18 & \textbf{$-$13.5}\\
\midrule
\multicolumn{4}{@{}l}{\textit{Temperature ($\tau$) = 0.6}}\\[1pt]
BF16 (value) & 90.2 & 6.65 & 100\\
\cmidrule(lr){1-4}
$\Delta$ MXFP4 & $-$8.5 & $-$1.06 & $-$11.5\\
$\Delta$ Per-step TempQuant, forward KL & $-$21.8 & $-$1.12 & $-$15.5\\
$\Delta$ Per-step TempQuant, JSD & $-$12.6 & $-$1.00 & $-$12.7\\
$\Delta$ Per-step TempQuant, reverse KL & \textbf{$-$5.5} & \textbf{$-$0.96} & \textbf{$-$10.7}\\
\midrule
\multicolumn{4}{@{}l}{\textit{Temperature ($\tau$) = 1.0}}\\[1pt]
BF16 (value) & 83.0 & 6.24 & 100\\
\cmidrule(lr){1-4}
$\Delta$ MXFP4 & $-$8.7 & $-$0.59 & $-$9.1\\
$\Delta$ Per-step TempQuant, forward KL & $-$15.8 & $-$0.67 & $-$11.4\\
$\Delta$ Per-step TempQuant, JSD & $-$11.0 & $-$0.62 & $-$9.8\\
$\Delta$ Per-step TempQuant, reverse KL & \textbf{$-$2.6} & \textbf{$-$0.56} & \textbf{$-$7.3}\\
\bottomrule
\end{tabular*}
\caption{MXFP4 version of Table~\ref{tab:xsum-diversity}: self-BLEU, how alike a policy's summaries of the same article are (lower is more varied), quality (ROUGE-2 against human summaries), and closeness to BF16 (Section~\ref{sec:downstream}). BF16 rows give its values, and $\Delta$ rows give each policy's change from BF16. Bold marks the closest policy.}
\label{tab:mxfp4-xsum}
\end{table}

\paragraph{Correctness agreement on MATH.}
Table~\ref{tab:math-correctness-flips} extends Figure~\ref{fig:correctness-flips}
to all three temperatures, on the 497 MATH questions whose reference answers
can be scored, counting both directions of disagreement with BF16. Two BF16
runs already disagree on about 67, 67, and 53 of 1,988 attempts at the three
temperatures, and the best loss removes $23\%$, $37\%$, and $24\%$ of the
disagreements MXFP4 adds on top of this.

\begin{table}[!htbp]
\centering\small
\setlength{\tabcolsep}{4pt}
\begin{tabular*}{\linewidth}{@{\extracolsep{\fill}}lrrrr@{}}
\toprule
Quantized policy & Accuracy (\%) & \shortstack{BF16 correct,\\quantized wrong}
 & \shortstack{BF16 wrong,\\quantized correct} & Total disagreements\\
\midrule
\multicolumn{5}{@{}l}{\emph{Temperature $\tau=0.3$}}\\
Native MXFP4 & 92.20 & 92 & \textbf{23} & 115\\
Forward KL & 92.73 & 88 & 29.5 & 117.5\\
Reverse KL & 92.40 & 88.5 & 23.5 & 112\\
JSD & 92.96 & 79 & 25 & \textbf{104}\\
Global TempQuant & \textbf{93.20} & \textbf{74} & 30 & \textbf{104}\\
\midrule
\multicolumn{5}{@{}l}{\emph{Temperature $\tau=0.6$}}\\
Native MXFP4 & 92.71 & 84 & 22 & 106\\
Forward KL & \textbf{93.36} & 75.5 & 26.5 & 102\\
Reverse KL & 93.08 & 74 & \textbf{19.5} & 93.5\\
JSD & 93.34 & \textbf{70.5} & 21 & \textbf{91.5}\\
Global TempQuant & 92.50 & 84 & 23 & 107\\
\midrule
\multicolumn{5}{@{}l}{\emph{Temperature $\tau=1.0$}}\\
Native MXFP4 & 92.81 & 81 & \textbf{21} & 102\\
Forward KL & \textbf{93.46} & \textbf{68.5} & 21.5 & \textbf{90}\\
Reverse KL & 93.41 & 72 & 24 & 96\\
JSD & 92.93 & 79 & 21.5 & 100.5\\
Global TempQuant & 92.45 & 87 & 25 & 112\\
\bottomrule
\end{tabular*}
\caption{MXFP4 MATH accuracy and correctness disagreements with BF16 on 1,988 scoreable attempts per temperature. Per-step rows average two fits. Bold marks the highest accuracy and the fewest disagreements at each temperature.}
\label{tab:math-correctness-flips}
\end{table}

\paragraph{Pooled test across tasks.}
\label{app:mxfp4-preservation-test}
Figure~\ref{fig:correctness-flips} shows one task at one temperature, where
there are few disagreements to count. To check whether the effect holds more
broadly, we combine all four reasoning tasks at all three temperatures into
one test, fixed before we computed it. For each question, we compare how
often a policy answers it correctly with how often BF16 does: if BF16 is
right on 3 of 4 samples and the policy on 1 of 4, they differ by $0.5$. We
average this difference over all questions, and compare native MXFP4 with the
JSD and reverse KL controllers. To decide whether a difference is larger than
chance, we resample the questions many times
(Appendix~\ref{app:behavior-protocol}), correcting for running two tests.
Table~\ref{tab:mxfp4-pooled-test} gives the result. For reference, two halves
of BF16's own samples differ by $0.081$ on this measure.

\begin{table}[!htbp]
\centering\small
\begin{tabular*}{\linewidth}{@{\extracolsep{\fill}}lrrrr@{}}
\toprule
Policy & Distance to BF16 & Reduction [95\% CI] & $p$ & Holm $p$\\
\midrule
Native MXFP4 & 0.1301 & & & \\
Per-step TempQuant, JSD & \textbf{0.1242} & \textbf{0.0059 [0.0013, 0.0106]} & \textbf{0.011} & \textbf{0.021}\\
Per-step TempQuant, reverse KL & 0.1303 & $-$0.0002 [$-$0.0054, 0.0051] & 0.94 & 0.94\\
Per-step TempQuant, forward KL & 0.1263 & 0.0038 [$-$0.0015, 0.0091] & 0.16 & \\
\bottomrule
\end{tabular*}
\caption{Average difference between each policy's and BF16's share of correct answers per question on MXFP4, over four tasks and three temperatures. The test compares the JSD and reverse KL controllers with native MXFP4; forward KL is shown for reference. Bold marks the policy closest to BF16.}
\label{tab:mxfp4-pooled-test}
\end{table}

\subsection{Qwen3-4B AWQ}
\label{app:awq}
We apply global TempQuant to Qwen's official 4-bit AWQ release of Qwen3-4B
(revision 74d4bd2, served in float16 with vLLM 0.33), which loses little
accuracy, and apply the forward KL per-step predictor trained on GPTQ3 to it
unchanged. Table~\ref{tab:awq-kl} gives the forward KL to BF16 on the
held-out steps at $\tau=0.6$, and Table~\ref{tab:awq} the accuracy over two
samples per question.

\begin{table}[!htbp]
\centering\small
\setlength{\tabcolsep}{4pt}
\begin{tabular*}{\linewidth}{@{\extracolsep{\fill}}lrr@{}}
\toprule
Setting for AWQ & Temperature & Forward KL\\
\midrule
Requested, $(0.6,.95)$ & 0.60 & 0.237\\
Top-$p$ off, $(0.6,1)$ & 0.60 & 0.063\\
AWQ's global TempQuant, $(0.71,1)$ & 0.71 & \textbf{0.061}\\
GPTQ3's global TempQuant, $(1.13,1)$ & 1.13 & 0.088\\
GPTQ3's per-step predictor & 1.7 (mean) & 0.148\\
\bottomrule
\end{tabular*}
\caption{Qwen3-4B AWQ forward KL to BF16 on held-out steps at $\tau=0.6$ under each setting.}
\label{tab:awq-kl}
\end{table}

\begin{table}[!htbp]
\centering\small
\setlength{\tabcolsep}{4pt}
\begin{tabular*}{\linewidth}{@{\extracolsep{\fill}}lrrrr@{}}
\toprule
Policy & GSM8K & MATH & GPQA & At length limit (\%)\\
\midrule
AWQ, $(0.6, .95)$ & \textbf{94.7} & \textbf{95.3} & 41.2 & 0.3\\
Global TempQuant, $(0.71, 1)$ & 94.5 & 94.7 & \textbf{45.2} & \textbf{0.1}\\
\bottomrule
\end{tabular*}
\caption{Accuracy (\%) of Qwen3-4B AWQ at $\tau=0.6$. Bold marks the better value in each column (lower for the length limit).}
\label{tab:awq}
\end{table}

\section{Divergence measures and oracle analysis}
\label{app:theory}
\subsection{Forward KL, reverse KL, and JSD}
\label{app:divergences}
With natural logarithms and $m=\frac{p+q}{2}$,
\begin{align}
 \KL(p\|q)&=\sum_{i:p_i>0}p_i\log\!\left(\frac{p_i}{q_i}\right),\nonumber\\
 D_{\mathrm{JS}}(p,q)&=\frac{1}{2}\KL(p\|m)+\frac{1}{2}\KL(q\|m),
 \label{eq:final-losses}
\end{align}
and reverse KL swaps $p$ and $q$. Forward KL is infinite when the quantized
model removes a token BF16 keeps, and reverse KL in the opposite case. JSD
is always between zero and $\log 2$.

\subsection{KL with smoothed probabilities}
\label{app:smoothed-divergences}
To keep KL finite, we mix a small uniform distribution $U$ into both
distributions, $P_\eta=(1-\eta)P+\eta U$ with $\eta=10^{-4}$, and compute
\begin{equation}
 D_{F,\eta}(P,Q)=\KL(P_\eta\|Q_\eta),\qquad
 D_{R,\eta}(P,Q)=\KL(Q_\eta\|P_\eta).
 \label{eq:smoothed-kl}
\end{equation}
If the quantized model removes probability mass $M$ that BF16 keeps, the
penalty grows like $M\log(1/\eta)$ as $\eta$ shrinks. Training scales both
KL losses by $1/4$; reported values are unscaled.

\subsection{Temperature-only oracle}
\label{app:temperature-oracle}
For a target $p$ and $q_\beta=\softmax(\beta z)$ with $\beta=1/T$ and
$\mu=\sum_i p_i z_i$,
\begin{equation}
 L(\beta)=\log\sum_i e^{\beta z_i}-\beta\mu-H(p),\qquad
 L'(\beta)=\E_{q_\beta}[z]-\mu,\qquad
 L''(\beta)=\Var_{q_\beta}(z).
 \label{eq:final-scalar}
\end{equation}
The problem is convex in $\beta$, and at an interior optimum
$\E_{q_\beta}[z]=\E_p[z]$ and $H(q^*)=H(p)+\KL(p\|q^*)$, so the best
temperature does not match entropy unless the distributions match
exactly \citep{mattei2026welltempered}.

On the full distributions, without top-$p$, the best per-step temperature
removes $73\%$, $53\%$, and $33\%$ of GPTQ3's forward KL at requested
temperatures $0.3$, $0.6$, and $1.0$.

\subsection{Forward-KL oracle for temperature and \texorpdfstring{top-$p$}{top-p}}
\label{app:joint-oracle}
Table~\ref{tab:fitted-control-summary} compares the oracle with both forms
of TempQuant and their smaller versions.
Per-step TempQuant's held-out forward KL is above global TempQuant's
because its learned cutoff sometimes removes tokens BF16 keeps, which
forward KL penalizes heavily; with top-$p$ turned off, the two match.
Allowing the oracle any cutoff, down to keeping only the top token, lowers
its forward KL by about another 40\% at $0.6$.

\begin{table}[!htbp]
\centering
\setlength{\tabcolsep}{6pt}
\renewcommand{\arraystretch}{1.15}
\begin{tabular*}{\linewidth}{@{\extracolsep{\fill}}lrrr@{}}
\toprule
Policy & $\tau=0.3$ & $\tau=0.6$ & $\tau=1.0$\\
\midrule
Uncorrected & 2.409 & 1.493 & 0.867\\
Global TempQuant (top-$p$ only) & 1.037 & 0.548 & 0.364\\
Global TempQuant ($T$ only) & 0.564 & 0.513 & 0.474\\
Global TempQuant & 0.451 & 0.397 & 0.346\\
\midrule
Oracle per-step TempQuant ($T$ only) & 0.288 & 0.264 & 0.255\\
Oracle per-step TempQuant & 0.272 & 0.240 & 0.220\\
Per-step TempQuant ($T$ only) & 0.738 & 0.510 & 0.494\\
Per-step TempQuant & 0.594 & 0.470 & 0.484\\
Per-step TempQuant, top-$p$ off & 0.449 & 0.400 & 0.344\\
\midrule
Oracle per-step TempQuant: share removed (\%) & 89 & 84 & 75\\
\bottomrule
\end{tabular*}
\caption{GPTQ3 smoothed forward KL to BF16 in nats. The oracle sees BF16 at every token and picks the best setting from the grid global TempQuant searches. Per-step TempQuant, top-$p$ off, keeps the predicted temperature and turns top-$p$ off. Per-step values average five training runs, except that row, which averages the two runs evaluated downstream; the two averages differ by under $3\%$.}
\label{tab:fitted-control-summary}
\end{table}

For unsmoothed forward KL, the oracle must keep every token BF16 gives
positive probability, and keeping more tokens only adds to the loss. With
logits sorted so $z_1\geq\cdots\geq z_V$ and $k^*=\max\{i:p_i>0\}$, the
joint problem reduces to minimizing
\[
 L_{k^*}(\beta)=\log\sum_{j\leq k^*}e^{\beta z_j}
            -\beta\sum_i p_i z_i-H(p)
\]
over the temperature range and then picking a cutoff that keeps exactly
the top $k^*$ tokens. This is again convex in $\beta$.

\subsection{The oracle in generation}
\label{app:downstream-oracle}
At every step, the oracle chooses the temperature and top-$p$ that minimize
forward KL to BF16 and samples from GPTQ3's own filtered distribution. All
policies run with the same seeds and code on the same GPU for each question.
On MATH, the oracle is ahead of per-step TempQuant by $21.9$ points
[$10.9$, $32.8$] and of global TempQuant by $25.0$ [$14.1$, $35.9$].
This gap shows how much a per-step predictor could still gain by better
recognizing, from the quantized model alone, when the top token has flipped,
which we leave to future work.

\begin{table}[!htbp]
\centering\small
\setlength{\tabcolsep}{4pt}
\begin{tabular*}{\linewidth}{@{\extracolsep{\fill}}lrrrrrr@{}}
\toprule
 & \multicolumn{3}{c}{GSM8K (96 questions)} & \multicolumn{3}{c}{MATH (64 questions)}\\
\cmidrule(lr){2-4}\cmidrule(l){5-7}
Policy & Accuracy & Length (K) & At limit & Accuracy & Length (K) & At limit\\
\midrule
BF16, same questions & 96.6 & 2.3 & 0.3 & 93.4 & 4.9 & 1.2\\
\cmidrule(lr){1-7}
GPTQ3 & 47.9 & 12.2 & 63.5 & 25.0 & 29.8 & 89.1\\
Global TempQuant & 84.4 & 7.6 & 19.8 & 53.1 & 19.0 & 42.2\\
Per-step TempQuant & 79.7 & 8.3 & 25.5 & 56.3 & 20.4 & 46.9\\
\textbf{Oracle per-step TempQuant} & \textbf{92.7} & \textbf{4.8} & \textbf{3.1} & \textbf{78.1} & \textbf{13.3} & \textbf{21.9}\\
\bottomrule
\end{tabular*}
\caption{GPTQ3 on the first GSM8K and MATH questions at $\tau=0.6$ (accuracy and responses at the length limit in \%). Per-step TempQuant pools its two training runs; BF16 averages four samples from the original runs.}
\label{tab:downstream-oracle}
\end{table}

\subsection{Limits of correcting token rankings}
\label{app:mechanism-development}
A positive temperature keeps the logit ranking, and top-$p$ keeps a leading
set without reordering it, so a lower-ranked token can never become the most
likely one. If BF16's top token has rank $r$ in the quantized model, any
setting gives it probability at most $1/r$.
GPTQ3 keeps BF16's most likely token at $83\%$ of steps and MXFP4 at
$89\%$. When GPTQ3 changes it, BF16's token is usually second, which happens
at $10\%$ of steps, and it falls below fifth at only $1.8\%$.
Table~\ref{tab:flip-share} splits forward KL by these steps at each
requested temperature, before and after the oracle. Table~\ref{tab:residual}
splits it further by the rank GPTQ3 gives BF16's most likely token and, when
that rank is 1, by whether GPTQ3 keeps the order of the tokens BF16 keeps
after top-$p$.
The largest share of the error sits where BF16's top token falls to second
place. This is the most common kind of flip, at about $10\%$ of steps, and
it holds $37$--$41\%$ of the error the oracle leaves at every requested
temperature, more than any other group.

Figure~\ref{fig:which-way} uses the training and held-out steps together at
$\tau=0.6$, and fits the temperature only, with top-$p$ off.

\begin{table}[!htbp]
\centering\small
\setlength{\tabcolsep}{4pt}
\begin{tabular*}{\linewidth}{@{\extracolsep{\fill}}lrrrrrr@{}}
\toprule
 & \multicolumn{2}{c}{Top token kept (83\%)} & \multicolumn{2}{c}{Top token changed (17\%)} & \multicolumn{2}{c}{Share on changed steps (\%)}\\
\cmidrule(lr){2-3}\cmidrule(lr){4-5}\cmidrule(l){6-7}
$\tau$ & GPTQ3 & Oracle & GPTQ3 & Oracle & GPTQ3 & Oracle\\
\midrule
0.3 & 0.28 & 0.03 & 12.66 & 1.45 & 90 & 92\\
0.6 & 0.35 & 0.04 & 7.01 & 1.22 & 81 & 87\\
1.0 & 0.31 & 0.05 & 3.55 & 1.02 & 70 & 80\\
\bottomrule
\end{tabular*}
\caption{Forward KL per step from GPTQ3 to BF16 on 768 held-out steps, split by whether GPTQ3 keeps BF16's most likely token, before and after the per-step oracle, and the share of the total on steps where it changes.}
\label{tab:flip-share}
\label{tab:rank-flips}
\end{table}

\begin{table}[!htbp]
\centering\small
\setlength{\tabcolsep}{4pt}
\begin{tabular*}{\linewidth}{@{\extracolsep{\fill}}lrrrr@{}}
\toprule
Steps grouped by GPTQ3 rank of BF16 top token & \shortstack{Steps\\(\%)} & \shortstack{Raw forward\\KL (\%)} & \shortstack{Oracle\\residual (\%)} & \shortstack{Removed by\\oracle (\%)}\\
\midrule
\multicolumn{5}{@{}l}{\textit{Temperature ($\tau$) = 0.3; oracle removes 89\% overall}}\\[1pt]
Top token kept, order kept & 80.1 & 5.1 & 1.3 & 97\\
Top token kept, order changed & 3.3 & 3.7 & 6.0 & 82\\
BF16 top ranked 2nd & 9.9 & 44.4 & 41.2 & 90\\
BF16 top ranked 3rd--5th & 4.9 & 33.6 & 33.3 & 89\\
BF16 top ranked below 5th & 1.8 & 13.3 & 18.1 & 85\\
\midrule
\multicolumn{5}{@{}l}{\textit{Temperature ($\tau$) = 0.6; oracle removes 84\% overall}}\\[1pt]
Top token kept, order kept & 78.0 & 8.8 & 1.8 & 97\\
Top token kept, order changed & 5.3 & 9.5 & 10.3 & 83\\
BF16 top ranked 2nd & 9.9 & 30.7 & 39.7 & 79\\
BF16 top ranked 3rd--5th & 4.9 & 33.0 & 31.4 & 85\\
BF16 top ranked below 5th & 1.8 & 18.1 & 16.9 & 85\\
\midrule
\multicolumn{5}{@{}l}{\textit{Temperature ($\tau$) = 1.0; oracle removes 75\% overall}}\\[1pt]
Top token kept, order kept & 73.0 & 11.1 & 2.7 & 94\\
Top token kept, order changed & 10.3 & 18.3 & 16.9 & 77\\
BF16 top ranked 2nd & 9.9 & 28.3 & 37.3 & 67\\
BF16 top ranked 3rd--5th & 4.9 & 26.6 & 28.1 & 73\\
BF16 top ranked below 5th & 1.8 & 15.7 & 14.9 & 76\\
\bottomrule
\end{tabular*}
\caption{Where the error left by the per-step oracle's temperature and top-$p$ sits, for GPTQ3 on 768 held-out steps, grouped by the rank GPTQ3 gives BF16's most likely token (\%).}
\label{tab:residual}
\end{table}

\section{Runtime cost}

\label{app:completion-results}

\label{app:runtime-timing}
Table~\ref{tab:decode-timing} measures the cost of each generated token
during decoding. Global TempQuant only changes the two numbers passed to the
sampler, so it costs nothing, and turning top-$p$ off makes it slightly
faster. Per-step TempQuant adds $0.5$--$2$ ms, or $18$--$34\%$, per token, and
peak memory is the same for all three. The predictor itself is small; most
of this time goes to passes over the full vocabulary to compute its
features and apply top-$p$, which our implementation does not optimize and
which fused kernels could shorten. The added time grows with batch size
(Table~\ref{tab:decode-timing}). We time MXFP4; GPTQ3 is served as
dequantized BF16 weights, so the predictor adds the same work per token.

\begin{table}[!htbp]
\centering\small
\setlength{\tabcolsep}{6pt}
\begin{tabular*}{\linewidth}{@{\extracolsep{\fill}}lrrrr@{}}
\toprule
Batch size & Requested settings & Global TempQuant & Per-step TempQuant & Per-step increase\\
\midrule
1 & 2.93 & 2.89 & 3.47 & 18\%\\
8 & 3.48 & 3.41 & 4.64 & 33\%\\
64 & 6.09 & 5.89 & 8.19 & 34\%\\
\bottomrule
\end{tabular*}
\caption{Milliseconds per generated token for MXFP4 Qwen3-4B in vLLM on one AMD Instinct MI350X GPU, with 512 prompt tokens and 1,024 generated tokens, averaged over three repeats.}
\label{tab:decode-timing}
\end{table}

\section{Use of AI tools}
\label{app:ai-assistance}
Generative AI assisted with planning, literature review, code and tests,
derivations, writing, and figures. LLM coding agents were also used extensively
to run and monitor experiments, collect model outputs, and process and analyze
the resulting data.

We also tried Sakana AI's ShinkaEvolve\footnote{\url{https://github.com/SakanaAI/ShinkaEvolve}},
which evolves code with LLM-proposed changes, to search for improvements to
earlier temperature predictors, scoring each candidate by its held-out
divergence from BF16. None improved clearly enough over the baseline to be
adopted in the final method.

The authors remain responsible for the experiments, claims, and manuscript.
Automated checks do not replace human verification.

\end{document}